\documentclass[twocolumn]{article}

\usepackage[numbers, sort&compress]{natbib}

\usepackage[utf8]{inputenc} %
\usepackage[T1]{fontenc}    %
\usepackage{hyperref}       %
\usepackage[]{authblk}
\usepackage{url}            %
\usepackage{booktabs}       %
\usepackage{amsfonts}       %
\usepackage{nicefrac}       %
\usepackage{microtype}      %
\usepackage{algorithmic}
\usepackage{algorithm}
\usepackage{amsmath}
\usepackage{amssymb}
\usepackage{bm}
\usepackage{todonotes}

\newcommand{\includescaled}[2][0.8]{\includegraphics[width=#1\textwidth,height=#1\textheight,keepaspectratio]{#2}}

\DeclareMathOperator{\Tr}{Tr}
\newcommand{\trp}{{^\top}} %
\newcommand{\inv}{^{-1}}
\newcommand{\W}{\mathbf{W}}
\newcommand{\Cb}{\mathbf{C}}
\newcommand{\D}{\mathbf{D}}
\newcommand{\E}{\mathbf{E}}
\newcommand{\A}{\mathbf{A}}
\newcommand{\R}{\mathbf{R}}
\newcommand{\Sb}{\mathbf{S}}
\newcommand{\B}{\mathbf{B}}
\newcommand{\Q}{\mathbf{Q}}

\newcommand{\vecop}{\mathrm{vec}}
\newcommand{\I}{\mathbf{I}}
\newcommand{\X}{\mathbf{X}}
\newcommand{\M}{\mathbf{M}}
\newcommand{\Y}{\mathbf{Y}}

\DeclareMathOperator{\expect}{\mathbb{E}}
\renewcommand{\d}{\mathrm{d}}

\newcommand{\ew}{\mathbb{E}}

\renewcommand{\vec}[1]{\mathbf{#1}}
\newcommand{\bi}{\vec{b}\vec{1}\trp}

\title{Matrix-normal models for fMRI analysis}
   
\author[1]{Michael Shvartsman\thanks{Corresponding author: ms44@princeton.edu}}
\author[2]{Narayanan Sundaram}
\author[1]{Mikio C.\ Aoi}
\author[1]{Adam Charles}
\author[2]{Theodore L.\ Wilke}
\author[1]{Jonathan D.\ Cohen}
\affil[1]{ Princeton Neuroscience Institute, Princeton University}
\affil[2]{Parallel Computing Lab, Intel Corporation}

\begin{document} 

\maketitle

\begin{abstract}
Multivariate analysis of fMRI data has benefited substantially from advances in machine learning. Most recently, a range of probabilistic latent variable models applied to fMRI data have been successful in a variety of tasks, including identifying similarity patterns in neural data (Representational Similarity Analysis and its empirical Bayes variant, RSA and BRSA; Intersubject Functional Connectivity, ISFC), combining multi-subject datasets (Shared Response Mapping; SRM), and mapping between brain and behavior (Simultaneous Modeling). Although these methods share some underpinnings, they have been developed as distinct methods, with distinct algorithms and software tools. We show how the matrix-variate normal (MN) formalism can unify some of these methods into a single framework. In doing so, we gain the ability to reuse noise modeling assumptions, algorithms, and code across models. Our primary theoretical contribution shows how some of these methods can be written as instantiations of the same model, allowing us to generalize them to flexibly modeling structured noise covariances. Our formalism permits novel model variants and improved estimation strategies: in contrast to SRM, the number of parameters for MN-SRM does not scale with the number of voxels or subjects; in contrast to BRSA, the number of parameters for MN-RSA scales additively rather than multiplicatively in the number of voxels. We empirically demonstrate advantages of two new methods derived in the formalism: for MN-RSA, we show up to 10x improvement in runtime, up to 6x improvement in RMSE, and more conservative behavior under the null. For MN-SRM, our method grants a modest improvement to out-of-sample reconstruction while relaxing an orthonormality constraint of SRM. We also provide a software prototyping tool for MN models that can flexibly reuse noise covariance assumptions and algorithms across models.

\end{abstract}

\section{Introduction}

Functional magnetic resonance imaging (fMRI) analysis is a challenging problem for statistics and machine learning: signal-to-noise ratio for extracting scientifically meaningful information is low, 
and physiological and instrumentation noise creates correlations in space and time that can mask signal and magnify false alarms.
Recent methods have been developed in the statistics and machine learning community to address these challenges, including for dimension reduction and subject-to-subject mapping (SRM, \citep{Chen2015}; TFA, \citep{Manning2014}), estimation of patterns of neural similarity within and across subjects (BRSA, \citep{Cai2016}; ISFC, \citep{Simony2016}), and mapping from brain to behavior via latent cognitive models \citep[SM,][]{Turner2016}.

These models are similar in that they seek a latent, typically low-rank, structure in fMRI data using multivariate gaussian models. Yet they are different in the quantity they attempt to estimate, and in the estimation methods they use. 
Furthermore, these techniques are restricted to modeling only either temporal or spatial correlation (or neither), even though both spatial and temporal noise structure exists in the data. 
These differences make it difficult to share insights and advances across techniques. 
In this work we show that matrix-variate (MN) normal models provide a powerful formalism for understanding and developing fMRI data analysis methods in a unified way. 

Specifically, we show that many existing methods can be derived from the MN framework. MN variants of these methods are not restricted in their noise model and can simultaneously capture spatial and temporal noise. 
Furthermore, the shared mathematical structure enables the creation of an MN development software framework that admits flexible swapping between various covariance models---a task that otherwise involves substantial engineering effort. 

 %

Our contributions are as follows:
\begin{enumerate}
\item Motivation for MN models as a unifying mathematical model for fMRI analysis, illustrating its wide applicability with examples from both regression (via RSA) and factor analysis (via SRM).\item A toolkit for developing MN models using Tensorflow \cite{tensorflow2015-whitepaper}, and implementations of RSA and SRM variants that can model both spatial and temporal covariance. (\S\ref{sec:implementation}). %
\item An expectation-conditional-maximization (ECM)algorithm for fitting MN-based SRM, in which the number of parameters does not scale with the number of subjects (unlike conventional SRM), and the orthonormality constraint on the shared space projection is removed  (\S\ref{sec:implementation}). 
\item Demonstration that MN-based RSA is approximately an order of magnitude faster than the previous state of the art method, can be up to 6x more accurate (relative to synthetic ground truth) at SNRs as low as 0.08 and thousands of voxels, and is most conservative under the null hypothesis.
\item Demonstration that MN-based SRM can improve on SRM performance in terms of reconstruction (\S\ref{sec:results}).
\end{enumerate}

The remainder of the paper is organized as follows: we discuss background and related work in \S\ref{sec:background}. \S\ref{sec:math} provides motivation for our formalism, and derives RSA and SRM in this framework, and the ECM algorithm for MN-SRM. 
\S\ref{sec:implementation} discusses our software implementation and the challenges involved therein. We show the results of our experiments in \S\ref{sec:results} and conclude in \S\ref{sec:conclusion} with some discussion as to how other cutting edge analyses fall into our framework. 

\section{Background}
\label{sec:background}
fMRI uses the magnetic properties of oxygenated blood to measure blood flow in the brain as a proxy for neural computation. fMRI data exhibits temporal and spatial correlations due to blood flow dynamics, acquisition constraints, and the spatially distributed temporally evolving mental computation itself. 
With computational and theoretical advances, Multi-Voxel Pattern Analysis (MVPA; \citealp{Norman2006}) has leveraged successes in machine learning for decoding more sophisticated representations and processes from fMRI data. A number of recent analyses pipelines have relied on gaussian latent variable models due to their wide applicability and computational tractability. We focus on two here -- SRM and RSA -- due to their wide use and broad applicability, though we treat two additional models in the supplementary material. 
\subsection{Representational Similarity Analysis (RSA)}

The goal of RSA \citep{Kriegeskorte2008} is to use distances between correlations or other (dis)similarity metrics between responses to stimuli in the fMRI dataset to theoretically predicted distances. This approach has been particularly successful in the visual domain \citep[e.g.][]{Yamins2013,Haxby2011}. Due to the isomorphism between correlation and regression, it has been shown that the standard RSA estimator is biased (in a formal sense) when applied to within-run data \citep{Cai2016}, but the state of the art empirical Bayes method based on maximum marginal likelihood (BRSA) mitigates this particular bias \citep{Cai2016}. 

\subsection{Shared response mapping}

A challenge in analyzing grouped data (e.g.\ coming from multiple subjects) is that while we expect subjects to have similar mental responses to a given stimuli, these responses may be idiosyncratically realized in the neural signal. This is essentially the repeated measures problem, seen here in a discriminative perspective. 
For classification-based decoding and other discriminative analyses that fall under the rubric of MVPA, managing this is
called the hyperalignment problem. Hyperalignment models project all subjects into a shared space that is used for analysis. SRM \citep{Chen2015} is one recent hyperalignment method that is a factor analytic model, which linearly projects all subjects' data into a shared, low-dimensional functional timecourse. SRM can be used for feature selection to enable state of the art decoding performance.\section{Matrix Normal Models for fMRI} 
\label{sec:math}

Conventional multivariate fMRI analysis methods choose whether to model noise covariance in space or in time, while assuming independence in the other dimension, an assumption violated in real fMRI data as noted above. 
\begin{figure}[t]
\includescaled[0.5]{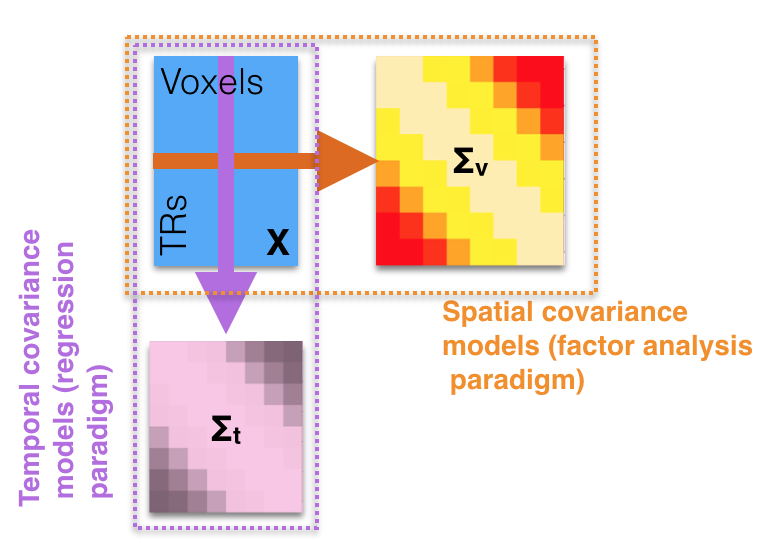}
\label{fig:motivation}
\vspace{-0.2in}
\caption{\textbf{Matrix normal models simultaneously model spatial and temporal noise.}}
\end{figure}

MN models, also known as kronecker-separable covariance models, provide a formalism addressing the problem of multivariate data analysis \citep[e.g.][]{Bijma2005,Werner2008}. The matrix-normal distribution is defined as:
\begin{align}
\X \sim & \mathcal{MN}_{mn}(\M,\R,\mathbf{C})\\
\log p(&\X\mid \M,\R, \mathbf{C}) = -2\log mn - m \log|\mathbf{C}|
 \\
 &- n \log|\R| - \Tr\left[\mathbf{C}\inv(\X-\M)\trp\R\inv(\X-\M)\right]. \nonumber
\end{align}

The intuition behind the kronecker separability is that if $\Y\sim\mathcal{MN}(\M, \R, \Cb)$ then $\vecop(\Y)\sim\mathcal{N}(\vecop(\M),\Cb\otimes\R)$, where $\otimes$ is the kronecker product operator and $\vecop$ is the vectorization operator. 
In the case of fMRI, a kronecker-separable covariance assumes that spatial covariance is the same at every time, and temporal covariance is the same for every voxel (see Fig.~\ref{fig:motivation} for an illustration of this point). 
The covariance between any two voxels at two times is a product of their space and time covariance. 

In this section, we show how the popular representational similarity analysis (RSA; \citealp{Kriegeskorte2008,Cai2016}) and shared response mapping (SRM; \citealp{Chen2015})  methods in neuroimaging can be written as matrix-normal models. We perform similar derivations for intersubject functional connectivity (ISFC; \citealp{Simony2016}) and Simultaneous modeling (SM; \citealp{Turner2016}) in the supplement. We begin with RSA \citep{Kriegeskorte2008}. Standard correlation-based RSA estimates stimulus-by-stimulus distances in brain activity space. If the distance matrix is a correlation matrix, this process is equivalent to encoding the predicted process model components (e.g.\ Markov states for reinforcement learning, or neural network activations) in a design matrix $\X$, under the linear model \citep{Cai2016}:
\begin{align}
\vec{y}_i\mid\X,\beta_i,\tau^2 &\sim \mathcal{N}(\X\beta_i,\tau^{-2}\I),
\label{eq:multivariate-rsa}
\end{align}

where $\tau^{-2}$ is the residual precision, $\vec{y}_i$ is the (centered) timecourse of the $i$th voxel, and the coefficient vector $\beta_i$ is the response pattern of each voxel to the modeled stimulus. The empirical row correlation of the $\beta$'s is the RSA correlation matrix. 
If one uses point estimates of $\beta$ to compute the RSA correlation matrix
the estimator is biased, and this bias can inject structure from the design matrix into the estimate \citep{Cai2016}. Bayesian RSA (BRSA; \citep{Cai2016}) instead marginalizes over $\beta$:
\begin{align}
\beta\sim\mathcal{N}(0, \mathbf{U})\\
\vec{y}_i\mid\X,\beta_i,\tau^2 &\sim \mathcal{N}(0,\tau^{-2}\I + \X\mathbf{U}\X\trp),
\end{align}
and performs MAP estimation on $\mathbf{U}$ by gradient descent, mitigating this bias. 
Now we derive matrix-variate RSA. 
To write a matrix-normal density for RSA (MN-RSA), we stack all of the $\vec{y}_i$ vectors into a matrix $\Y$, and stack all of the regression weights $\beta$ into a matrix $\W$: 
\begin{align}
\Y\mid\W,\Sigma_t,\Sigma_v\sim\mathcal{MN}(\X\W, \Sigma_t, \Sigma_{v,\Y})\\
\W\mid\mathbf{U},\Sigma_v\sim\mathcal{MN}(0,\mathbf{U}, \Sigma_{v,\W})
\end{align}

This model can no longer model voxel-specific temporal correlations, but in return can model the residual spatial covariance $\Sigma_v$. This tradeoff will play out differently in different datasets. In this multilinear regression form, this problem appears similar to a number of models used recently in the multi-task learning literature \citep[e.g.][]{Bonilla2008,Skolidis2011,Stegle2011,Rakitsch2013,Greenewald2015}. However, unlike those settings, the estimation target in RSA is the covariance $\mathbf{U}$ rather than predicted data in new tasks.

In previous work on estimating kronecker-separable covariances, both the signal and noise spatial covariances are assumed to be different, i.e.\  $\Sigma_{v,\Y}\ne \Sigma_{v,\W}$. As a result, the marginal covariance (marginalizing over $\W$) is a sum of kronecker factors, which previous work had estimated using Permuted Rank-penalized Least Squares \citep{Greenewald2015,Greenewald2013,Greenewald2014} or gradient descent exploiting the compatibility between diagonalization and the kronecker product for efficient likelihood computation \citep{Stegle2011,Rakitsch2013}. In the domain of fMRI, spatial covariance are driven primarily by physiological factors (blood flow) and instrument constraints, both of which theoretically affect the signal and spatial noise covariances in an identical way. Consequently, we assume $\Sigma_v := \Sigma_{v,\Y} = \Sigma_{v,\W}$. This gives us a convenient matrix-normal marginal likelihood (see supplement for derivation): 
\begin{align}
\Y\mid\mathbf{U},\Sigma_t,\Sigma_v  \sim\mathcal{MN}(0, \Sigma_t+ \X\mathbf{U}\X\trp, \Sigma_v), 
\end{align}
which we term the \emph{MN-RSA} model.

\subsection{Matrix-variate shared response model}
Consider the following factor analysis model for fMRI data for multiple subjects: 
\begin{align}
\vec{y_{jk}}\mid\W_k,\vec{s}_j,\Sigma_v &\sim \mathcal{N}(\W_k\vec{s}_j,\tau_k^{-2}\Sigma_v),
\end{align}
where $\vec{y}_{jk}$ is a mean-centered vector containing all voxel activities at a single timepoint (rather than $\vec{y}_i$, the single voxel's time series in Eq.~\ref{eq:multivariate-rsa}). We have also added indexing by subject $k$, since SRM (unlike RSA) is a multi-subject method. $\vec{s}_j$ is a latent spatial map \emph{for all subjects} for that particular time point, and $\W$ is subject-specific a projection matrix from the shared map into that subject's data. $\Sigma_v$ is a shared \emph{spatial} noise covariance as in MN-RSA above, scaled by a subject-specific precision $\tau_k^{-2}$. To make a matrix-variate factor model, we row-stack $\vec{y}_{jk}\trp$ into $\Y_k\trp$, stack $\vec{s}_j$ into $\Sb$, and obtain the following model:
\begin{align}
\Y_k\trp\mid\W_k,\Sb,\mu,\Sigma_v,\sigma_k &\sim \mathcal{MN}_{v,n}(\W_k\Sb\trp,\tau^{-2}_k\Sigma_v,\Sigma_t)
\end{align}

This factor analysis model now has the exact same form as the regression model above, except that $\X_k$ is observed and $\Sb$ is latent. Both of these matrix-variate models now have the exact same form: a mean that is an intercept plus a product of two matrices, one fully-specified covariance, and an identity covariance. 

We can drop the subject indices by row-stacking all of the subject timecourses and weights and introducing a subject covariance $\rho:=\mathrm{diag}(\tau_1^{-2}, \tau_1^{-2}, \ldots, \tau_n^{-2})$: 
\begin{align}
\Sb &\sim\mathcal{N}(0, \Sigma_s, \Sigma_t)\\
\W &\sim\mathcal{N}(0, \rho\otimes\Sigma_v, \Sigma_w)\\
\mathbb{X}\mid\mathbb{W},\Sb,\Sigma_t,\Sigma_v,\rho &\sim \mathcal{MN}(\mathbb{W}\Sb, \rho\otimes \Sigma_v, \Sigma_t), 
\end{align}
giving us the \emph{MN-SRM} model\footnote{we omit the spatial mean $\mu_j$ in the derivation for brevity, though not in the implementation}. 
The covariances $\Sigma_w,\Sigma_s$ are both set to $\I$ to regularize the model. The model implies that all subjects share a latent time-course $\Sb$, as well as temporal and spatial noise covariances $\Sigma_v, \Sigma_t$ that are scaled independently for each subject\footnote{Since a kronecker-structured covariance is determined only up to a constant, the scale on $\Sigma_t$ and $\Sigma_v$ is isomorphic, except that by scaling $\Sigma_v$ we make the remainder of the derivation more straightforward.}. 

If $\Sigma_t=\Sigma_v=\I$ and $\W_k\trp\W_k=\I\quad \forall k$, this MN-SRM model is exactly the SRM model. However, in the MN formulation, we see that we have two marginalization choices: the first is marginalizing over the shared time-course $\Sb$, as SRM does, and the second is marginalizing over the mappings $\W$ instead. The latter marginal density estimates $vnk$ parameters ($v$ voxels, $n$ subjects, $k$ features) instead of $tk$ parameters ($t$ timepoints, $k$ features), which is appealing because for whole-brain analyses $v\gg t$. It also replaces the strong orthonormal constraint on $\W$ with a weaker Gaussian prior. This is a theoretically desirable property for the following reason: if the true data is not generated with orthonormal $\W$ per subject, forcing an orthonormal $\W$ makes $\Sb$ counter-rotate against it. With a single subject $\W\trp\W=\I$ w.l.o.g., but with multiple subjects, the best $\Sb$ for each subject is rotated differently to maintain orthonormality for that subject, giving a worse group $\Sb$. We later validate this intuition empirically in our reconstruction task. 

\section{Estimation and the \texttt{matnormal} prototyping tool} \label{sec:implementation}

We leverage the shared structure of MN models to develop a unified framework for estimation using Python and the Tensorflow library\footnote{all code will be made available on github with a standard \texttt{scikit-learn} API}. The implementation is flexible in the specification of noise covariances: for a (spatial or temporal) covariance $\Sigma$, the API only requires implementations of $\Sigma^{-1}\X$ and $\log|\Sigma|$ given $\X$ for efficient computation of marginal likelihoods marginalizing in either the row or column direction. This gives users the ability to choose the noise model complexity relative to the size of their data -- or the ability to explore a large number of models with simple noise quickly before selecting a more complex noise model for later analysis. 

All other required routines can be derived from these, including marginalization that automatically leverages efficiencies derived from non-marginal covariance structures using the Woodbury and Sylvester lemmas. The amount of shared code allows, for example, MN-RSA to be implemented in only 50 lines of python code. We have implemented isotropic, diagonal, full rank, AR(1) kernel, squared exponential kernel, and Kronecker factored covariances. That is, we can further factor the spatial covariance $\Sigma_v$ into $\Sigma_z\otimes\Sigma_y\otimes\Sigma_x$, where $x,y,z$ are the spatial dimensions. Using kronecker-factored spatial covariances in fMRI is challenging because, since the brain is not a perfect cube, the masking of voxels that do not contain brain partially violates the kronecker structure. We address this challenge by developing a fast algorithm for the inverse and determinant of a masked kronecker-factored covariance (detailed in the supplement), increasing the toolbox's utility to a wider variety of fMRI datasets.
We also include example implementations of MN (multilinear) regression, MN factor analysis MN-RSA, and MN-SRM. 

Using this toolbox, we can perform maximum marginal likelihood estimation using gradient descent, leveraging gradients automatically computed with Tensorflow and our covariance API for rapid prototyping. In practice, this is sufficient for single-subject estimates, and our results for MN-RSA below are all using gradient descent. Directly maximizing the marginal likelihood for group models such as MN-SRM is substantially slower because RSA is fit to a single subject whereas SRM is fit to a ten subjects or more, an order of magnitude more data. 

To mitigate this issue, we derive an efficient expectation conditional maximization (ECM) algorithm for learning MN-SRM, which estimates the sufficient statistics of $\Sb$ in the E-step and performs conditional maximization updates of the remaining parameters in the M-step. When estimating MN-SRM using ECM in our toolkit, we can still impose structure on $\Sigma_v$ and $\Sigma_t$, however only certain constraints allow closed-form covariance updates. Due to space constraints we delegate the ECM derivation to the supplementary material. The algorithm does not exploit any special properties of MN-SRM relative to other MN models, and should be applicable to them with minor changes. 

\section{Results} \label{sec:results}

We validate the MN framework for fMRI by exploring the behavior of MN-RSA and MN-SRM in simulations and real data. 
To demonstrate the accuracy of MN-RSA, we explore its performance on synthetic data. Our focus on synthetic data is because neither out-of-sample prediction nor real-data ground truth for RSA is well-defined in the literature, with the standard measure for evaluating RSA methods comparing the estimated covariance to a behaviorally relevant matrix. Since RSA matrices consistent with behavior can arise due to estimator bias alone \citep{Cai2016}, this metric is not useful. We do show MN-RSA performance on real data to verify that it does not recover spurious correlations when the design matrix and brain data are unrelated, and highlight the need for the field to develop better predictive validation metrics for covariance estimation. 

For MN-SRM, we perform two experiments. The first is an out-of-sample reconstruction experiment testing whether the shared response we recover can reconstruct a new subject's data. The second is using MN-SRM for feature extraction with the goal of classification. 
\subsection{MN-RSA}

For the MN-RSA experiment we compared BRSA~\citep{Cai2016} against MN-RSA with diagonal spatial covariance temporal covariance consisting of an AR(1) component, plus a low-rank matrix with the rank set to 15. For the spatial variance, the intuition is that by learning the variance of each voxel we can better tune SNR. %
For temporal covariance, AR(1) is simple, expressive and comparable to BRSA. We excluded naive RSA from this comparison because of its known bias, and since BRSA has been shown to achieve superior performance \citep{Cai2016}. \begin{figure}[h]
  \begin{center}
  \includescaled[0.4]{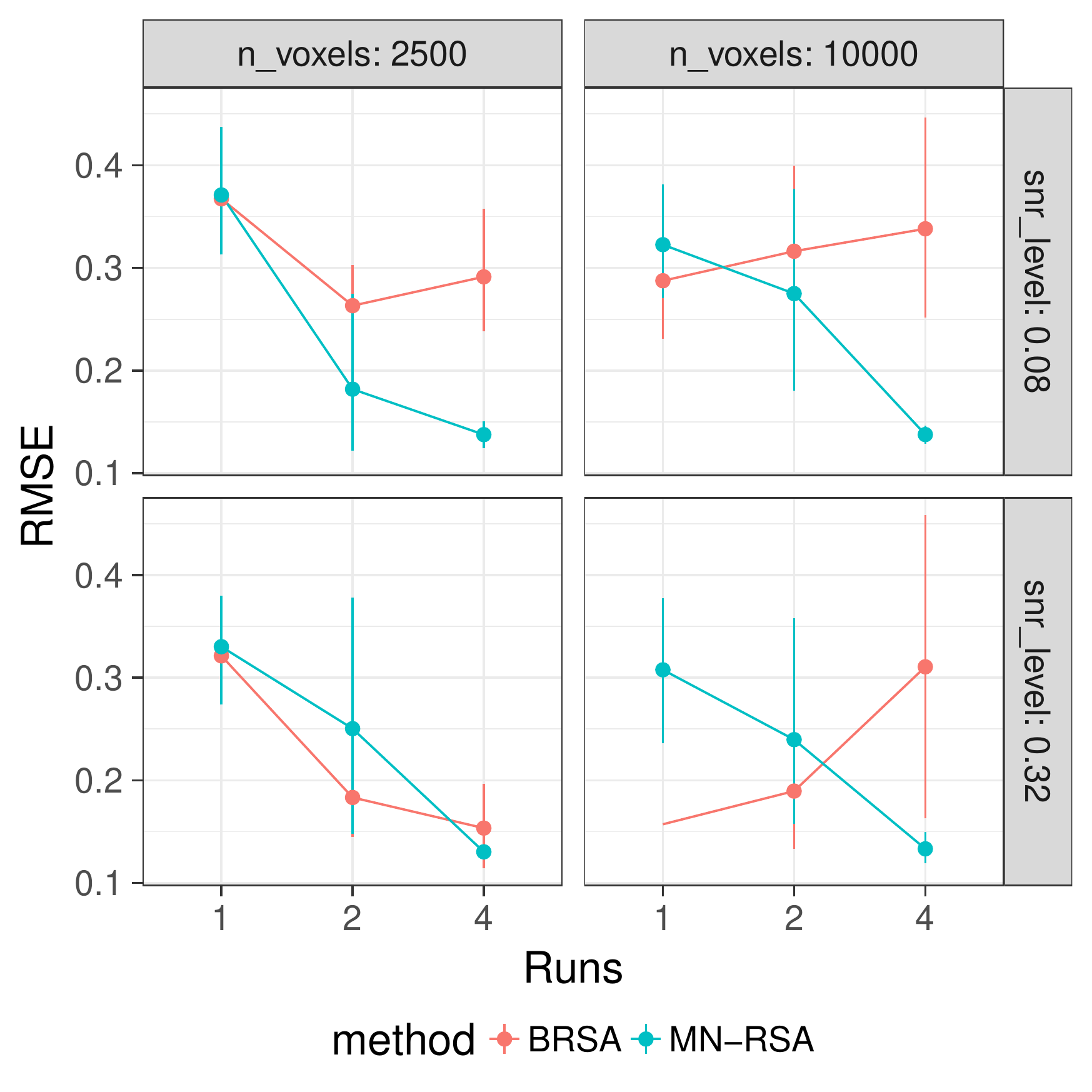}
  \end{center}
  \vspace{-0.2in}
  \caption{\textbf{MN-RSA performs better at larger numbers of voxels and lower SNR.} For smaller datasets (e.g.\ 400 voxels; not shown) and larger SNRs, BRSA performs better. The improved performance of MN-RSA is enabled by not modeling temporal noise independently for each voxel.}
\label{fig:brsa-synth}
\end{figure}

\paragraph{Experiment 1: synthetic data}
We generated synthetic data using the BRSA example in the \texttt{brainiak} package. This synthetic dataset has AR(1) noise in the temporal domain, spatial noise generated from a gaussian process, and a number of design-irrelevant timecourses included. Thus, it violates both models' assumptions, and includes spatial structure that is challenging for non-matrix-variate models to handle. The synthetic datasets included two different SNR levels, three different numbers of timepoints known in the field as `TRs' (equivalent to 1, 2, and 4 runs of the experiment), and two different numbers of voxels (2500, 10000) to show how the algorithms scale with noise, time, and space. We replicated each combination 10 times. Each model was run and timed separately on a full node of a compute cluster with two Intel$^{\mbox{\tiny\textregistered}}$ Xeon$^{\mbox{\tiny\textregistered}}$ E5-2670 processors at 2.6 GHz with hyper-threading enabled. The deviation from ceiling performance on simple synthetic training data suggests that these methods are not overfitting. MN-RSA is up to 10x faster than the reference implementation of BRSA (estimated using BFGS) on the largest problems (figure in supplement). 

Fig.~\ref{fig:brsa-synth} shows estimated root-mean-squared-error (RMSE) performance against the true correlation matrix for the different numbers of voxels, TRs, and SNRs. 
MN-RSA obtains lower RMSE than BRSA in most settings, with the difference being particularly stark at larger problem sizes. 
 In addition, because it estimates fewer parameters, MN-RSA can be up to 10x faster on the same hardware for large-scale problems. \paragraph{Experiment 2: null data}
As noted above, there is no ground truth evaluation or RSA, and has been shown to recover spurious results under the null hypothesis when there is structure in the design matrix (for conventional RSA) or model mismatch in noise covariance (for all RSA methods) \citep{Cai2016}. Consequently, an important evaluation of RSA methods is their behavior under the null hypothesis, i.e.\ where no signal is expected to exist. 

In this experiment, we use a real resting state dataset \citep{VanEssen2012} in which subjects are not given a task, with a random temporally contiguous window of 186 TRs selected from each participant and a lateral occipital cortex region of interest (ROI). With this resting state dataset, we used the same design matrix as in the experiment above, which is completely unrelated to the dataset. We show two example subjects' RSA covariance matrices under all three methods in Fig.~\ref{fig:rsa-null}, and the remainder in the supplement. Since MN-RSA estimates both the low-rank temporal structure and the $\mathbf{U}$ matrix simultaneously, it is capable of assigning next to zero variance to $\mathbf{U}$ if the design matrix is unrelated to the data. 
This feature means that for most subjects under the null hypothesis, $\mathbf{U}$ correctly approaches 0, and the RSA correlation matrix is clearly degenerate, in contrast to RSA and BRSA, both of which produce the appearance of structure. In the supplement, we also show the distribution of the elements of the estimated RSA matrix for each subject, with a clear spike at zero in a majority of subjects only for MN-RSA, showing that it is the most conservative method under the null hypothesis. 

\begin{figure*}[h]
\includescaled[0.9]{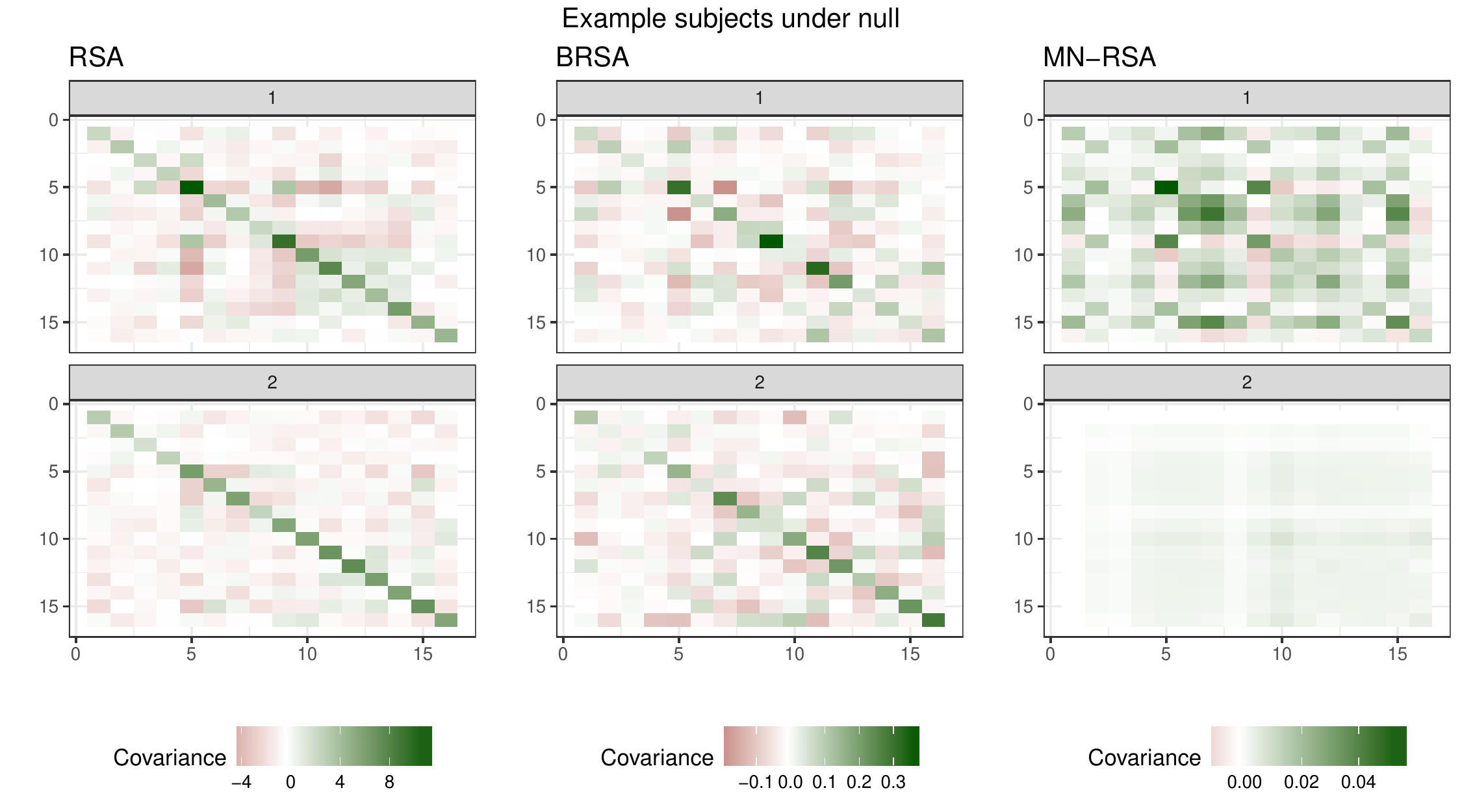}
\label{fig:rsa-null}
\caption{\textbf{MN-RSA is the only method that delivers obviously degenerate results under the null.} Estimates for all but three of the subjects look like subject 2 for MN-RSA, BRSA estimates all look homogeneous, and naive RSA estimates are about evenly split between finding low rank structure like subject 1 and not finding it like subject 2.}
\end{figure*}

\subsection{MN-SRM}

We test two variants of MN-SRM. The first sets $\Sigma_v=\Sigma_t=\I$, differing from SRM only in the marginalization direction and the removal of the orthonormality constraint on $\W$. Since it has the same relationship to SRM that dual probabilistic PCA \citep{Lawrence2005} has to PCA, we call it dual probabilistic SRM (DP-SRM). The second variant, MN-SRM, uses the diagonal $\Sigma_v$ and AR(1) $\Sigma_t$ we used for MN-RSA, above. We compare these models to SRM, as well as to ICA as a naive baseline. For these experiments, we use the \emph{raider} \citep{Haxby2011} and \emph{sherlock} \citep{Chen2016} datasets (see Tab.~\ref{tab:datasets}for detail). 

\begin{table*}[]
\begin{tabular}{@{}llllr@{}}
\toprule
\textbf{Dataset} & \textbf{Subjs.} & \textbf{TRs} & \textbf{Region of Interest}       & \textbf{Voxels} \\ \midrule
sherlock \citep{Chen2016}         & 16                & 1976         & Posterior Medial Cortex           & 813             \\
raider \citep{Haxby2011}          & 10                & 2203         & Ventral Temporal Cortex & 1000            \\ 
HCP \citep{VanEssen2012}           & 29                & 186    & Lateral Occipital Cortex & 2000            \\ 
\end{tabular}
\caption{fMRI dataset properties used for experiments 2, 3, and 4. We thank the authors for sharing their data.}
\label{tab:datasets}
\end{table*}

\vspace{-0.15in}
\paragraph{Experiment 3: out of sample reconstruction}

To test each model's ability to recover the shared latent time-course, we perform a held-out reconstruction experiment. We fit the factorization methods on all but one subject with 10, 30, or 50 features, and then learn a projection from the shared time-course into that new subject. Our loss metric is the reconstruction error of the held out subject's data using the estimated shared time-course, and the new subject's map. In both, we use the portion of the dataset where subjects watched the same movie (Raiders of the Lost Ark, and an episode of BBC's Sherlock). 

\begin{figure*}[t]
\includescaled[1]{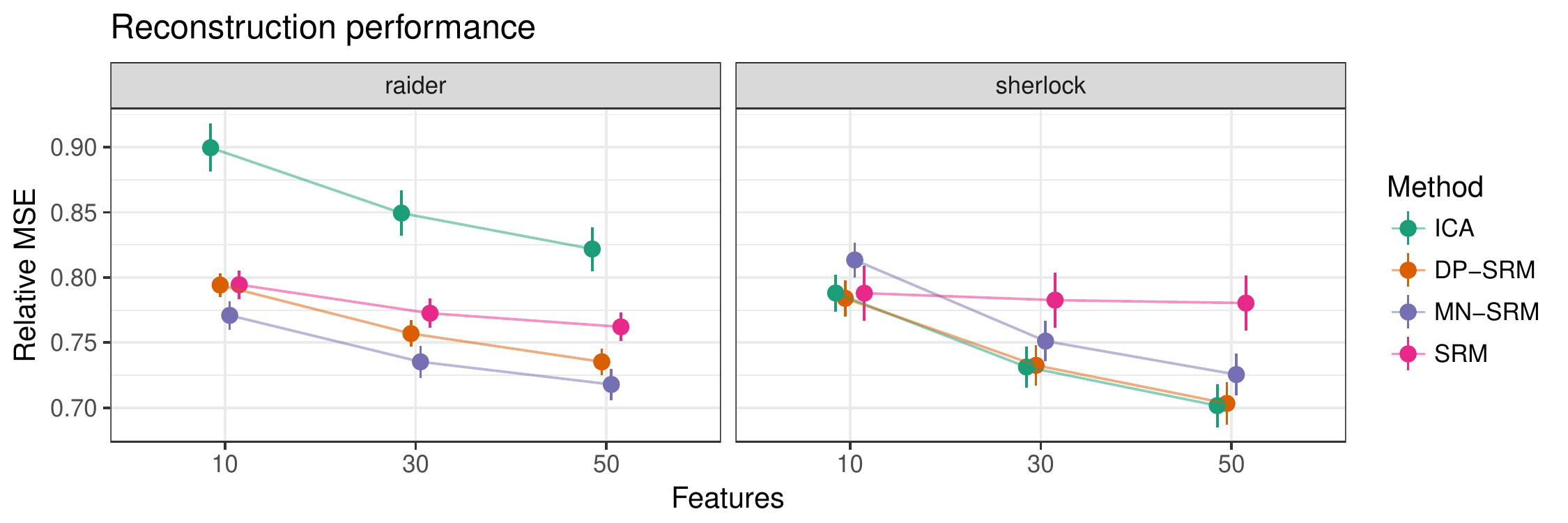}\\
\vspace{-0.25in}
\caption{\textbf{MN-SRM and DP-SRM reconstruct the same or better than SRM and ICA.} All models are trained on n-1 subjects, and the shared timecourse used to reconstruct the $n$th subject. Plotted are means and standard error across subjects.}
\label{fig:srm-reconstruction}
\end{figure*}

Fig.~\ref{fig:srm-reconstruction} shows that the reconstruction error of both MN methods is consistently lower than that of SRM in the \emph{raider} dataset, and lower in all but the smallest numbers of features in the \emph{sherlock} dataset. The improvement of MN methods over SRM validates our assertion that we should be able to more effectively fit our model by marginalizing over a larger number of parameters, and shows that benefit of MN models' flexibility in removing the orthonormality constraint on $\W$. The relative performance between the MN methods on the two datasets is also interesting: on \emph{raider}, adding the noise covariance modeling improves performance, whereas on \emph{sherlock} it does not. The ultimate reason for this is an interesting scientific question, and provides validation for our approach of flexible noise covariance modeling: there may not be a one-size-fits-all hyperalignment method. 

\paragraph{Experiment 4: feature extraction for classification}
One of the primary use-cases for SRM is as feature extraction method for classification. For this reason, and because classification performance was previously used to compare hyperalignment methods \citep{Chen2015}, we report performance on this task next. In both \emph{raider} and \emph{sherlock}, subjects begin by watching a movie clip, and then perform a cognitive task. SRM and similar methods can be used to learn a projection into a shared space while subjects are watching the same movie stimulus, and then use that learned mapping to project fMRI data recorded during the cognitive task. For \emph{raider} this task was viewing one of 7 possible images, and for \emph{sherlock} it was free-recalling scenes in the movie. We train a linear SVM to discriminate between the images subjects viewed in \emph{raider}, and between the scenes in \emph{sherlock}. 

\begin{figure*}[t]
\includescaled[1]{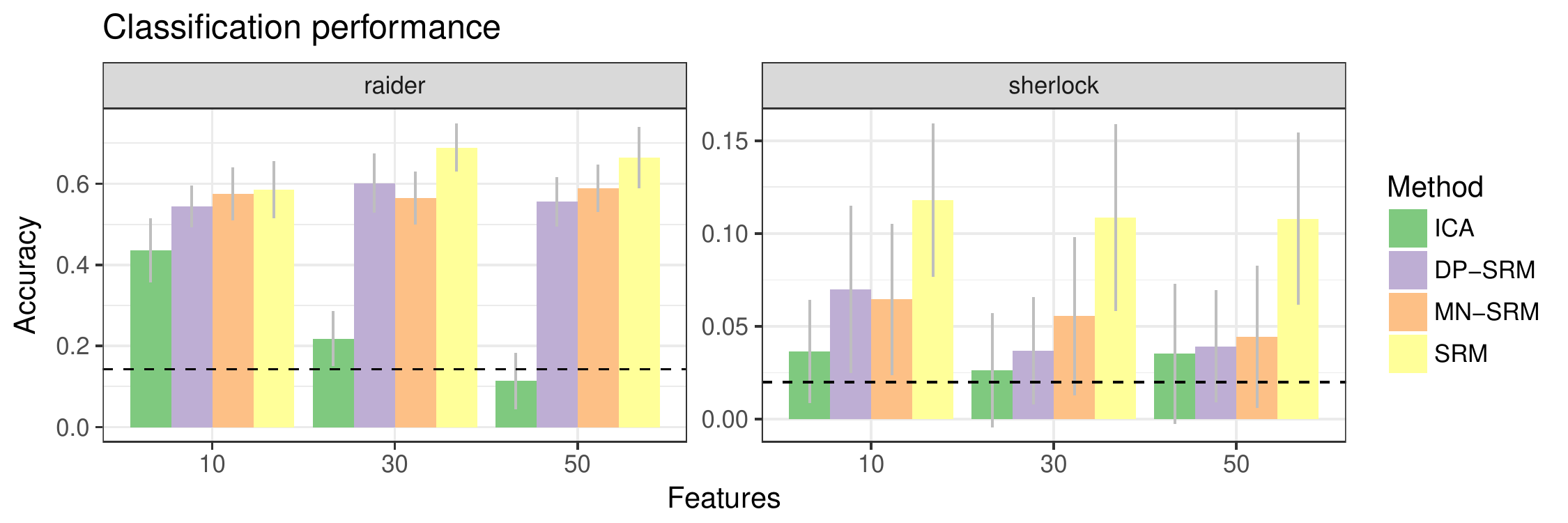}
\vspace{-0.25in}
\caption{\textbf{MN-SRM and DP-SRM approach SRM performance in feature extraction, while relaxing the orthonormality constraint on $\W$.} We train the SRM on all subjects watching a movie, and project the other task data into a shared space for classification. Plotted are means and error bars of out-of-sample prediction across subjects. The dashed line is chance performance.}
\vspace{-0.15in}
\label{fig:srm-classification}
\end{figure*}

Fig~\ref{fig:srm-classification} shows that in spite of our methods' better performance on the reconstruction task, their ability to extract features useful for linear classification lags behind the original SRM method. We suspect that this discrepancy between reconstruction and feature extraction points to a role for the orthonormality constraint as a regularizer, discarding variance shared across subjects that is specific to the movie data as opposed to the task data. \section{Discussion and conclusion} \label{sec:conclusion}

Probabilistic multivariate analyses of fMRI data are a promising direction of research, combining the interpretability previously associated with univariate analyses with the power of multivariate approaches. However, advances tend to proceed independently of each other, with distinct methods and algorithms for different problems. At face value this is not surprising as they have substantial differences: SRM and TFA are unsupervised, while BRSA and ISFC are supervised; BRSA and ISFC are somewhat unusual in seeking the correlation matrices of latent variables, whereas SRM is more conventionally concerned with latent space projection, and TFA with inferring brain networks. In addition, they all use distinct fitting techniques: gradient-based maximum marginal likelihood for BRSA, expectation-maximization for SRM, and variational inference for TFA.

We showed how some of these methods can be viewed as closely related matrix-variate models, and how the matrix-variate view allows us to simultaneously model spatial and temporal noise covariances in both methods. 
In neuroscience, such models have been applied to MEG/EEG data \citep{Ros2014}, as well as non-latent models for fMRI data \citep{Hartvig2002}, with some evidence that a separable covariance is a reasonable approximation to fMRI data even though voxel temporal correlations vary with spatial location. Our work contrasts with this previous work both in its unification of distinct methods, and in bringing matrix-variate latent variable models to fMRI analysis more broadly. 

In the MN view, we can show the relationship of some supervised fMRI analysis methods (RSA and ISFC) to multi-task regression and more broadly to kronecker-structured covariance models. Such models have been applied in areas as diverse as recommendation systems \citep{Allen2010}, environmental science \citep{Genton2007,Tsiligkaridis2013}, MIMO channel behavior \citep{Werner2007,Werner2008}, collaborative filtering \citep{Yu2009b}, compiler performance prediction and student test score modeling \citep{Bonilla2008}, video understanding \citep{Greenewald2014}, and genomics \citep{Yin2012,Rakitsch2013,Stegle2011}. However, in contrast to this existing work (and especially \citep{Stegle2011,Rakitsch2013}, which is closest to our contribution), the nature of fMRI noise admits simpler noise covariance assumptions that in turn yield different techniques for efficient likelihood computation, and a novel expectation-conditional-maximization algorithm. 

In addition to our theoretical contribution, we provided a software package for estimating the above models that allows for flexible assumptions about noise covariance, and provided evidence that for best performance, noise covariance assumptions may need to be adjusted for different datasets and tasks. 
Our experiments also revealed opportunity for future work. For example, MN-RSA performed worse than the previous method at larger numbers of TRs and smaller numbers of voxels (figure not shown), we suspect partially because of our method's inability to model different noise covariances for each voxel. 
Alternatively, it may exploit the connection between RSA and multi-task regression apparent in the matrix-variate formalization to bring techniques from multi-task regression to this latent covariance estimation problem. 
Likewise, while MN-SRM performed better at reconstruction than SRM, it did not produce features that improved classification performance. A broader exploration of noise models may help here, but we suspect that the true next gain may come from using the matrix-variate view to bring SRM and RSA even closer together into a unified formalism. Regardless, our toolkit will enable rapid prototyping as we progress in this domain. 

\clearpage
\bibliographystyle{abbrvnat_nourl_noissn}

\begin{thebibliography}{33}
\providecommand{\natexlab}[1]{#1}
\providecommand{\url}[1]{\texttt{#1}}
\expandafter\ifx\csname urlstyle\endcsname\relax
  \providecommand{\doi}[1]{doi: #1}\else
  \providecommand{\doi}{doi: \begingroup \urlstyle{rm}\Url}\fi

\bibitem[Abadi et~al.(2015)Abadi, Agarwal, Barham, Brevdo, Chen, Citro,
  Corrado, Davis, Dean, Devin, Ghemawat, Goodfellow, Harp, Irving, Isard, Jia,
  Jozefowicz, Kaiser, Kudlur, Levenberg, Man{\'{e}}, Monga, Moore, Murray,
  Olah, Schuster, Shlens, Steiner, Sutskever, Talwar, Tucker, Vanhoucke,
  Vasudevan, Vi{\'{e}}gas, Vinyals, Warden, Wattenberg, Wicke, Yu, and
  Zheng]{tensorflow2015-whitepaper}
M.~Abadi, A.~Agarwal, P.~Barham, E.~Brevdo, Z.~Chen, C.~Citro, G.~S. Corrado,
  A.~Davis, J.~Dean, M.~Devin, S.~Ghemawat, I.~Goodfellow, A.~Harp, G.~Irving,
  M.~Isard, Y.~Jia, R.~Jozefowicz, L.~Kaiser, M.~Kudlur, J.~Levenberg,
  D.~Man{\'{e}}, R.~Monga, S.~Moore, D.~Murray, C.~Olah, M.~Schuster,
  J.~Shlens, B.~Steiner, I.~Sutskever, K.~Talwar, P.~Tucker, V.~Vanhoucke,
  V.~Vasudevan, F.~Vi{\'{e}}gas, O.~Vinyals, P.~Warden, M.~Wattenberg,
  M.~Wicke, Y.~Yu, and X.~Zheng.
\newblock {{\{}TensorFlow{\}}: Large-Scale Machine Learning on Heterogeneous
  Systems}, 2015.

\bibitem[Allen and Tibshirani(2010)]{Allen2010}
G.~I. Allen and R.~Tibshirani.
\newblock {Transposable regularized covariance models with an application to
  missing data imputation}.
\newblock \emph{The Annals of Applied Statistics}, 4\penalty0 (2):\penalty0
  764--790, jun 2010.

\bibitem[Bijma et~al.(2005)Bijma, {De Munck}, and Heethaar]{Bijma2005}
F.~Bijma, J.~C. {De Munck}, and R.~M. Heethaar.
\newblock {The spatiotemporal MEG covariance matrix modeled as a sum of
  Kronecker products}.
\newblock \emph{NeuroImage}, 27\penalty0 (2):\penalty0 402--415, 2005.

\bibitem[Bonilla et~al.(2008)Bonilla, Chai, and Williams]{Bonilla2008}
E.~Bonilla, K.~M. Chai, and C.~Williams.
\newblock {Multi-task Gaussian Process Prediction}.
\newblock \emph{Nips}, 20\penalty0 (October):\penalty0 153--160, 2008.

\bibitem[Cai et~al.(2016)Cai, Schuck, Pillow, and Niv]{Cai2016}
M.~B. Cai, N.~W. Schuck, J.~W. Pillow, and Y.~Niv.
\newblock {A Bayesian method for reducing bias in neural representational
  similarity analysis}.
\newblock In \emph{NIPS Proceedings}, pages 1--10, sep 2016.

\bibitem[Chen et~al.(2016)Chen, Leong, Honey, Yong, Norman, and
  Hasson]{Chen2016}
J.~Chen, Y.~C. Leong, C.~J. Honey, C.~H. Yong, K.~A. Norman, and U.~Hasson.
\newblock {Shared memories reveal shared structure in neural activity across
  individuals}.
\newblock \emph{Nature Neuroscience}, 20\penalty0 (1):\penalty0 115--125, dec
  2016.

\bibitem[Chen et~al.(2015)Chen, Chen, Yeshurun, Hasson, Haxby, and
  Ramadge]{Chen2015}
P.-H.~C. Chen, J.~Chen, Y.~Yeshurun, U.~Hasson, J.~Haxby, and P.~J. Ramadge.
\newblock {A Reduced-Dimension fMRI Shared Response Model}.
\newblock \emph{Neural Information Processing Systems Conference (NIPS)}, pages
  460--468, 2015.

\bibitem[Genton(2007)]{Genton2007}
M.~G. Genton.
\newblock {Separable approximations of space-time covariance matrices}.
\newblock \emph{Environmetrics}, 18\penalty0 (7):\penalty0 681--695, nov 2007.

\bibitem[Greenewald and Hero(2015)]{Greenewald2015}
K.~Greenewald and A.~O. Hero.
\newblock {Robust Kronecker Product PCA for Spatio-Temporal Covariance
  Estimation}.
\newblock \emph{IEEE Transactions on Signal Processing}, 63\penalty0
  (23):\penalty0 6368--6378, dec 2015.

\bibitem[Greenewald et~al.(2013)Greenewald, Tsiligkaridis, and
  Hero]{Greenewald2013}
K.~Greenewald, T.~Tsiligkaridis, and A.~O. Hero.
\newblock {Kronecker sum decompositions of space-time data}.
\newblock In \emph{2013 5th IEEE International Workshop on Computational
  Advances in Multi-Sensor Adaptive Processing (CAMSAP)}, number~2, pages
  65--68. IEEE, dec 2013.

\bibitem[Greenewald and Hero(2014)]{Greenewald2014}
K.~H. Greenewald and A.~O. Hero.
\newblock {Kronecker PCA based spatio-temporal modeling of video for dismount
  classification}.
\newblock page 90930V, jun 2014.

\bibitem[Hartvig(2002)]{Hartvig2002}
N.~V. Hartvig.
\newblock {A Stochastic Geometry Model for Functional Magnetic Resonance
  Images}.
\newblock \emph{Scandinavian Journal of Statistics}, 29\penalty0 (3):\penalty0
  333--353, 2002.

\bibitem[Haxby et~al.(2011)Haxby, Guntupalli, Connolly, Halchenko, Conroy,
  Gobbini, Hanke, and Ramadge]{Haxby2011}
J.~V. Haxby, J.~S. Guntupalli, A.~C. Connolly, Y.~O. Halchenko, B.~R. Conroy,
  M.~I. Gobbini, M.~Hanke, and P.~J. Ramadge.
\newblock {A Common, High-Dimensional Model of the Representational Space in
  Human Ventral Temporal Cortex}.
\newblock \emph{Neuron}, 72\penalty0 (2):\penalty0 404--416, oct 2011.

\bibitem[Kriegeskorte(2008)]{Kriegeskorte2008}
N.~Kriegeskorte.
\newblock {Representational similarity analysis – connecting the branches of
  systems neuroscience}.
\newblock \emph{Frontiers in Systems Neuroscience}, 2\penalty0
  (November):\penalty0 4, 2008.

\bibitem[Lawrence(2005)]{Lawrence2005}
N.~D. Lawrence.
\newblock {Probabilistic non-linear Principal Component Analysis with Gaussian
  Process Latent Variable Models}.
\newblock \emph{Journal of Machine Learning Research}, 6:\penalty0 1783--1816,
  2005.

\bibitem[Manning et~al.(2014)Manning, Ranganath, Norman, and Blei]{Manning2014}
J.~R. Manning, R.~Ranganath, K.~A. Norman, and D.~M. Blei.
\newblock {Topographic factor analysis: A Bayesian model for inferring brain
  networks from neural data}.
\newblock \emph{PLoS ONE}, 9\penalty0 (5), 2014.

\bibitem[Norman et~al.(2006)Norman, Polyn, Detre, and Haxby]{Norman2006}
K.~A. Norman, S.~M. Polyn, G.~J. Detre, and J.~V. Haxby.
\newblock {Beyond mind-reading: multi-voxel pattern analysis of fMRI data}.
\newblock \emph{Trends in Cognitive Sciences}, 10\penalty0 (9):\penalty0
  424--430, 2006.

\bibitem[Rakitsch et~al.(2013)Rakitsch, Lippert, Borgwardt, and
  Stegle]{Rakitsch2013}
B.~Rakitsch, C.~Lippert, K.~Borgwardt, and O.~Stegle.
\newblock {It is all in the noise: Efficient multi-task Gaussian process
  inference with structured residuals}.
\newblock \emph{Advances in Neural Information Processing Systems}, pages
  1466--1474, 2013.

\bibitem[Ro{\'{s}} et~al.(2014)Ro{\'{s}}, Bijma, de~Gunst, and
  de~Munck]{Ros2014}
B.~Ro{\'{s}}, F.~Bijma, M.~de~Gunst, and J.~de~Munck.
\newblock {A three domain covariance framework for EEG/MEG data}.
\newblock \emph{NeuroImage}, 119:\penalty0 305--315, oct 2014.

\bibitem[Simony et~al.(2016)Simony, Honey, Chen, Lositsky, Yeshurun, Wiesel,
  and Hasson]{Simony2016}
E.~Simony, C.~J. Honey, J.~Chen, O.~Lositsky, Y.~Yeshurun, A.~Wiesel, and
  U.~Hasson.
\newblock {Dynamic reconfiguration of the default mode network during narrative
  comprehension}.
\newblock \emph{Nature Communications}, 7\penalty0 (May 2015):\penalty0 12141,
  jul 2016.

\bibitem[Skolidis and Sanguinetti(2011)]{Skolidis2011}
G.~Skolidis and G.~Sanguinetti.
\newblock {Bayesian Multitask Classification With Gaussian Process Priors}.
\newblock \emph{IEEE Transactions on Neural Networks}, 22\penalty0
  (12):\penalty0 2011--2021, dec 2011.

\bibitem[Stegle et~al.(2011)Stegle, Lippert, Mooij, Lawrence, and
  Borgwardt]{Stegle2011}
O.~Stegle, C.~Lippert, J.~Mooij, N.~D. Lawrence, and K.~Borgwardt.
\newblock {Efficient inference in matrix-variate Gaussian models with iid
  observation noise}.
\newblock \emph{Advances in Neural Information Processing Systems 24 (NIPS
  2011)}, pages 630--638, 2011.

\bibitem[Tsiligkaridis and Hero(2013)]{Tsiligkaridis2013}
T.~Tsiligkaridis and A.~O. Hero.
\newblock {Covariance Estimation in High Dimensions Via Kronecker Product
  Expansions}.
\newblock \emph{IEEE Transactions on Signal Processing}, 61\penalty0
  (21):\penalty0 5347--5360, nov 2013.

\bibitem[Turner et~al.(2013)Turner, Forstmann, Wagenmakers, Brown, Sederberg,
  and Steyvers]{Turner2013}
B.~M. Turner, B.~U. Forstmann, E.-J. Wagenmakers, S.~D. Brown, P.~B. Sederberg,
  and M.~Steyvers.
\newblock {A Bayesian framework for simultaneously modeling neural and
  behavioral data.}
\newblock \emph{NeuroImage}, 72:\penalty0 193--206, may 2013.

\bibitem[Turner et~al.(2014)Turner, Sederberg, and McClelland]{Turner2014}
B.~M. Turner, P.~B. Sederberg, and J.~L. McClelland.
\newblock {Bayesian analysis of simulation-based models}.
\newblock \emph{Journal of Mathematical Psychology}, 2014.

\bibitem[Turner et~al.(2015)Turner, van Maanen, and Forstmann]{Turner2015}
B.~M. Turner, L.~van Maanen, and B.~U. Forstmann.
\newblock {Informing cognitive abstractions through neuroimaging: The neural
  drift diffusion model.}
\newblock \emph{Psychological Review}, 122\penalty0 (2):\penalty0 312--336,
  2015.

\bibitem[Turner et~al.(2016)Turner, Rodriguez, Norcia, McClure, and
  Steyvers]{Turner2016}
B.~M. Turner, C.~A. Rodriguez, T.~M. Norcia, S.~M. McClure, and M.~Steyvers.
\newblock {Why more is better: Simultaneous modeling of EEG, fMRI, and
  behavioral data}.
\newblock \emph{NeuroImage}, 128:\penalty0 96--115, mar 2016.

\bibitem[{Van Essen} et~al.(2012){Van Essen}, Ugurbil, Auerbach, Barch,
  Behrens, Bucholz, Chang, Chen, Corbetta, Curtiss, {Della Penna}, Feinberg,
  Glasser, Harel, Heath, Larson-Prior, Marcus, Michalareas, Moeller,
  Oostenveld, Petersen, Prior, Schlaggar, Smith, Snyder, Xu, and
  Yacoub]{VanEssen2012}
D.~{Van Essen}, K.~Ugurbil, E.~Auerbach, D.~Barch, T.~Behrens, R.~Bucholz,
  A.~Chang, L.~Chen, M.~Corbetta, S.~Curtiss, S.~{Della Penna}, D.~Feinberg,
  M.~Glasser, N.~Harel, A.~Heath, L.~Larson-Prior, D.~Marcus, G.~Michalareas,
  S.~Moeller, R.~Oostenveld, S.~Petersen, F.~Prior, B.~Schlaggar, S.~Smith,
  A.~Snyder, J.~Xu, and E.~Yacoub.
\newblock {The Human Connectome Project: A data acquisition perspective}.
\newblock \emph{NeuroImage}, 62\penalty0 (4):\penalty0 2222--2231, oct 2012.

\bibitem[Werner and Jansson(2007)]{Werner2007}
K.~Werner and M.~Jansson.
\newblock {Estimation of kronecker structured channel covariances using
  training data}.
\newblock \emph{European Signal Processing Conference}, \penalty0
  (Eusipco):\penalty0 1201--1205, 2007.

\bibitem[Werner et~al.(2008)Werner, Jansson, and Stoica]{Werner2008}
K.~Werner, M.~Jansson, and P.~Stoica.
\newblock {On Estimation of Covariance Matrices With Kronecker Product
  Structure}.
\newblock \emph{IEEE Transactions on Signal Processing}, 56\penalty0
  (2):\penalty0 478--491, feb 2008.

\bibitem[Yamins et~al.(2013)Yamins, Hong, and Cadieu]{Yamins2013}
D.~L.~K. Yamins, H.~Hong, and C.~Cadieu.
\newblock {Hierarchical Modular Optimization of Convolutional Networks Achieves
  Representations Similar to Macaque IT and Human Ventral Stream}.
\newblock \emph{Advances in Neural Information Processing Systems}, \penalty0
  (October):\penalty0 1--9, 2013.

\bibitem[Yin and Li(2012)]{Yin2012}
J.~Yin and H.~Li.
\newblock {Model selection and estimation in the matrix normal graphical
  model}.
\newblock \emph{Journal of Multivariate Analysis}, 107:\penalty0 119--140, may
  2012.

\bibitem[Yu et~al.(2009)Yu, Lafferty, Zhu, and Gong]{Yu2009b}
K.~Yu, J.~Lafferty, S.~Zhu, and Y.~Gong.
\newblock {Large-scale collaborative prediction using a nonparametric random
  effects model}.
\newblock In \emph{Proceedings of the 26th Annual International Conference on
  Machine Learning - ICML '09}, pages 1--8, New York, New York, USA, 2009. ACM
  Press.

\end{thebibliography}
\onecolumn

\section{ Appendix A : Matrix-normal intersubject functional connectivity and simultaneous modeling}

Here we derive matrix-normal variants of two additional models from the literature, intersubject functional connectivity \citep{Simony2016}, and simultaneous modeling \citep{Turner2013,Turner2014,Turner2016}. 

\subsection{Matrix-normal intersubject functional connectivity}

The goal of the ISFC method is to estimate a ``shared stimulus-induced covariance matrix'' in fMRI data as a way to measure functional connectivity between brain regions while abstracting over subject-specific connectivity patterns and extracting only the patterns that are consistent across subjects. The intuition behind the method is simple: it computes pairwise correlations between each subject's patterns and averages them. To prove that the method is indeed free of subject-specific bias, Simony and colleagues frame their model in terms of a gaussian generative model. Here is this generative model, rewritten in the matrix-normal formalism: 
\begin{align}
\A\mid \Cb &\sim \mathcal{MN}(0, \Cb, \I)\\
\D_i \mid \sigma^2_{\D} &\sim \mathcal{MN}(0, \sigma^2_{\D} \I, \I)\\
\Sb &\sim \mathcal{MN}(0, \I, \I)\\
\E \mid \Q &\sim \mathcal{MN}(0, \Q, \I)\\
\X_i &= (\A + \D_i)\Sb + \E_i
\end{align}

The ``shared stimulus-induced covariance matrix'' that the method is intended to estimate is $\Cb$, the row covariance of the projection matrix into latent space. The somewhat redundant formulation is needed to motivate the closed-form estimator used in the original method. However, the formulation required for the closed-form estimator places severe restrictions on the projection matrix $\Sb$, both in terms of its rank (which must be full) and distribution (which is independent standard normal). We instead simplify the model and integrate out the projection. Let $\W_i=\A + \D_i$, and rewrite:
\begin{align}
\A \mid \Cb &\sim \mathcal{MN}(0, \Cb, \I)\\
\W_i \mid \A, \sigma^2_{\D} &\sim \mathcal{MN}(\A, \sigma^2_{\D} \I, \I)\\
\Sb &\sim \mathcal{MN}(0, \I, \I)\\
\X_i\mid \W, \Sb, \Q &\sim \mathcal{MN}(\W_i\Sb, \Q, \I)
\end{align}

Then marginalize $\A$: 
\begin{align}
\W_i \mid \Cb, \sigma^2_{\D} &\sim \mathcal{MN}(0, \Cb+\sigma^2_{\D} \I, \I)\\
\Sb &\sim \mathcal{MN}(0, \I, \I)\\
\X_i\mid \W, \Sb, \Q &\sim \mathcal{MN}(\W_i\Sb, \Q, \I)
\end{align}

The resultant model is remarkably similar to MN-SRM: ISFC models the row (spatial) noise covariance as full-rank whereas MN-SRM models it as diagonal. MN-SRM models the shared response covariance as full-rank but ISFC models it as diagonal. Finally, and most importantly, MN-SRM models the projection into latent space as orthonormal whereas ISFC is specifically interested in its covariance (which MN-SRM can in fact estimate). 

\subsection{Matrix-normal simultaneous modeling}

The simultaneous modeling framework \citep{Turner2015} is organized around attempts to estimate the joint covariance of the vector $\{\psi_1, \psi_2, \ldots, \psi_p, \phi_i, \phi_2, \ldots, \phi_k\}$, which is a combined vector of cognitive model parameters $\psi$ and features extracted from fMRI signal $\phi$. As it is a broad framework, a number of specific instances have been provided, with specific cognitive models including accumulator models and signal detection theory models, and feature extraction mechanisms including ICA, PCA, and other methods. 

There are a number of challenges with the current formulation of simultaneous modeling that we address: first, while the formulation in terms of correlations between brain and behavior allows for intuitive interpretation, it makes it challenging to regularize the model, or place priors on brain-behavior relationships, except for the special case of complete independence. Second, by performing the feature extraction in an unsupervised way, there is no guarantee that the features extracted will be relevant to the behavior or cognitive model; on the other hand, applying the framework to whole-brain data is not generally tractable, as it involves estimating a sizable covariance matrix by MCMC. 

We show how matrix-normal simultaneous modeling can address all of these challenges. Since SM is a framework rather than one specific model, and no public implementation is available, we focus on a toy example to illustrate our contribution. We choose factor analysis as our factor model, leave the cognitive model unspecified for the derivation, which is applicable to any cognitive model, and any \emph{linear} factor model. 

Here is the graphical model for simultaneous modeling (omitting the conjugate prior on $\Sigma$): 
\begin{align}
h_i\mid\psi_i &\sim \mathrm{Cog}(\psi_i)\\
\phi_i &= g(b_i)\\
\begin{bmatrix}\phi_i\\\psi_i\end{bmatrix}&\mid\mu_{\phi}, \mu_{\psi}, \Sigma_{\phi}, \Sigma_{\psi}, R_{\psi,\phi}\sim \mathcal{N}\left(\begin{bmatrix}\mu_{\phi}\\\mu_{\psi}\end{bmatrix},  \begin{bmatrix}
\Sigma_{\psi} & R_{\psi,\phi}\\
R_{\psi,\phi}\trp & \Sigma_{\phi}\\
 \end{bmatrix}\right),
\end{align}
where $h_i$ is a vector of behavioral outcomes at time $i$, $\Sigma_{\psi}$ is the model parameter covariance matrix, $\Sigma_{\phi}$ is the brain feature covariance matrix, and $R$ is the off-diagonal component corresponding to the covariance of brain and behavior. This formulation, esepcially if $R$ is further decomposed into standard deviations and correlations, makes the parameter estimates directly interpretable as correlations between brain sources and cognitive model parameters. However, this formulation challenging to constrain and regularize, because any regularization must respect the positive-definiteness constraint on the full covariance. 

We rewrite the model instead as a regression problem, which enables us to regularize, or in fact marginalize over nuisance parameters altogether. To do this, we write the conditional distribution of brain features on cognitive model parameters using the properties of partitioned Gaussians: 
\begin{align}
\label{eqn:cog-scalar}
h_i\mid\psi_i &\sim \mathrm{Cog}(\psi_i)\\
\psi_i &\sim \mathcal{N}(\mu_{\psi}, \Sigma_{\psi})\\
\phi_i \mid \psi_i &\sim \mathcal{N}(\mu_{\phi} + R_{\psi,\phi}\Sigma_{\psi}^{-1}(\psi-\mu_{\psi}),  \Sigma_{\phi} - R_{\psi,\phi}\Sigma_{\psi}^{-1}R_{\psi,\phi}\trp).
\end{align}

This model is equivalent to a regression model with structured residuals, as follows: 
\begin{align}
\ell_0 &:= \mu_{\phi}- R_{\psi,\phi}\Sigma_{\psi}^{-1}\mu_{\psi},\\
\ell &:=  R_{\psi,\phi}\Sigma_{\psi}^{-1}\\
\Sigma_{\Phi s} &:= \Sigma_{\phi} - R_{\psi,\phi}\Sigma_{\psi}^{-1}R_{\psi,\phi}\trp\\ 
\label{eqn:phi-scalar}
\phi_i \mid \psi_i &\sim \mathcal{N}(\ell_0 + \ell\psi_i,  \Sigma_{\Phi s}).
\end{align}

Now we can stack the model into matrix-variate form, also adding a design matrix for observed stimulus features $\X$ and its coefficient matrix $\beta$: 
\begin{align} 
\label{eqn:cog-matrix}
&\mathbf{H}\mid\bm{\Psi} \sim \mathrm{Cog.}(\bm{\Psi},\Sb)\\
\label{eqn:phi-matrix}
&\mathbf{\Phi}\mid\bm{\beta},\bm{\ell},\bm{\Sigma}_{\phi\mid\psi},\mathbf{S},\mathbf{\Psi} \sim \mathcal{MN}( \mathbf{\Psi}\bm{\ell}  + \X\beta,\Sigma_{\Phi t}, \bm{\Sigma}_{\Phi s}).
\end{align}
Expr.~\ref{eqn:cog-matrix} is the stacked version of Expr.~\ref{eqn:cog-scalar}, and Expr~\ref{eqn:phi-matrix} is the stacked version of Expr.~\ref{eqn:phi-scalar}, with the stimulus regression added and the intercept absorbed into the design matrix and covariance parameters introduced as needed. We now add a matrix-variate gaussian factor model for brain feature extraction: 
\begin{align}
&\Y\trp\mid\mathbf{\Phi}, \mathbf{W},\bm{\Sigma}_s, \bm{\Sigma}_t \sim \mathcal{MN}(\mathbf{W}\mathbf{\Phi} , \bm{\Sigma}_s,\bm{\Sigma}_t).
\end{align}
The resultant analysis combines a multilinear regression model for $\bm{\Phi}$ and matrix-factor model for $\Y$. In this case, since we only need the latent factors $\bm{\Phi}$ to map to the cognitive parameters $\bm{\Psi}$, we can marginalize over the factor mapping $\W$ by introducing a gaussian prior. We can likewise introduce priors over $\beta$ and $\ell$ and marginalize those variables out, since we do not need the regression mapping for decoding cognitive parameters $\bm{\Psi}$. This gives us a direct model from brain behavior via latent cognitive parameters and a neural factor space: 
\begin{align}
&\W \sim \mathcal{MN}(0, \bm{\Sigma}_s, \mathbf{I})\\
&\bm{\beta} \sim \mathcal{MN}(0, \bm{\Sigma}_{\Phi s}, \mathbf{U})\\
&\bm{\ell} \sim \mathcal{MN}(0, \bm{\Sigma}_{\Phi s}, \mathbf{V})\\
&\mathbf{H}\mid\bm{\Psi} \sim \mathrm{Cog.}(\bm{\Psi},\X)\\
&\mathbf{\Phi}\mid\bm{\Sigma}_{\Phi s},\bm{\Sigma}_{\Phi t},\mathbf{X},\mathbf{\Psi} \sim \mathcal{MN}( 0, \bm{\Sigma}_{\Phi s},
\bm{\Sigma}_{\Phi t} + \X\trp\mathbf{U}\X + \bm{\Psi}\trp\mathbf{V}\bm{\Psi})\\
&\Y\trp\mid\mathbf{\Phi},\bm{\Sigma}_s,\bm{\Sigma}_t \sim \mathcal{MN}(0, \bm{\Sigma}_s,\bm{\Sigma_}t + \mathbf{\Phi}\trp\mathbf{\Phi})
\end{align}

Given this marginalization, both the latent neural factors and the latent cognitive parameters appear in the model only as their inner products, and are not identifiable directly. Therefore, an equivalent model is a direct regression from voxels to cognitive parameters, marginalized over the mapping. This will be true for any linear factor model under this marginalization, giving the following final model: 
\begin{align}
&\mathbf{H}\mid\bm{\Psi}, \X \sim \mathrm{Cog.}(\bm{\Psi},\X)\\
&\Y\mid\bm{\Sigma}_{\Phi s},\bm{\Sigma}_{\Phi t},\mathbf{S},\mathbf{\Psi} ,\mathbf{U},\mathbf{V}\sim \mathcal{MN}( 0, \bm{\Sigma}_{s},\bm{\Sigma}_{t} + \X\mathbf{U}\trp\X + \bm{\Psi}\trp\mathbf{V}\bm{\Psi})
\end{align}

In this view we have arrived again at an RSA-type intuition, namely that while it may very challenging to know the true projection from $\Y$ to $\bm{\Psi}$, mapping them on second-order statistics in time space can prove to be useful, especially as the dimensionality of $\Y$ (and hence $\W$) grows. 

With the mapping marginalized, we can still perform prediction from the model by maximizing the likelihood of the cognitive parameters corresponding to new data given parameters estimated previously: 
\begin{align}
&\Y_{new} \mid \bm{\Psi} \sim \mathcal{MN}(\M, \hat{\bm{\Sigma}}_s, \Cb)\\
&\M = \Y_{old}(\hat{\Sigma}_{t} + \hat{\bm{\Psi}}\trp\hat{\bm{\Psi}})\inv(\hat{\bm{\Psi}}\trp\bm{\Psi})\\
&\Cb = \hat{\bm{\Sigma}}_t+ \bm{\Psi}\trp\bm{\Psi} -(\bm{\Psi}\trp\hat{\bm{\Psi}})(\hat{\Sigma}_{t} + \hat{\bm{\Psi}}\trp\hat{\bm{\Psi}})\inv(\hat{\bm{\Psi}}\trp\bm{\Psi}), 
\end{align}
where the hat-matrices are estimated previously and the remaining parameters are for new timepoints. This maximization rotates the inner-products of the old and new sets into the same orientation. If the old and new sets have different numbers of timepoints, we need to additionally replace the temporal noise covariance matrix with a kernel function, but otherwise the derivation proceeds identically. 

The resultant matrix-normal model mitigates the issues we identified previously: first, the only thing that scales with the number of voxels is the noise model rather than the mapping itself, allowing analysis to proceed using voxels directly assuming the noise model is efficient enough; second, it is targeted in that it automatically identifies the voxels most related to the cognitive model parameters; third, it is implicitly regularized via priors on $\beta$ and $\ell$. As with all MN models, it can also simultaneously handle both spatial and temporal noise in the fMRI signal. 

\section{ Appendix B : Derivation of matrix normal identities}

Consider the following three distributions: 

\begin{align}
\mathbf{X}_{ij} \sim \mathcal{MN}(\mathbf{A}_{ij}, \Sigma_{\mathbf{X}i},\Sigma_{\mathbf{X}j})\\
\mathbf{Y}_{jk} \sim \mathcal{MN}(\mathbf{B}_{jk}, \Sigma_{\mathbf{Y}j},\Sigma_{\mathbf{Y}k})\\
\mathbf{Z}_{ik}\mid\mathbf{X}_{ij},\mathbf{Y}_{jk} \sim \mathcal{MN}(\mathbf{X}_{ij}\mathbf{Y}_{jk} + \mathbf{C}_{ik}, \Sigma_{\mathbf{Z}_i}, \Sigma_{\mathbf{Z}_k})
\end{align}

We use lowercase subscripts to denote sizes, to make dimension constraints clearer. We first use the relationship between the matrix-normal and multivariate normal distribution to rewrite the densities in vectorized form. Next, we rewrite the vectorized product in the mean into kronecker form: 

\begin{align}
\vecop(\mathbf{Z}_{ik})\mid\mathbf{X}_{ij},\mathbf{Y}_{jk} \sim \mathcal{N}(\vecop(\X_{ij}\mathbf{Y}_{jk}+\mathbf{C}_{ik}), \Sigma_{\mathbf{Z}_k}\otimes\Sigma_{\mathbf{Z}_i})\\
\vecop(\mathbf{Z}_{ik})\mid\mathbf{X}_{ij},\mathbf{Y}_{jk} \sim \mathcal{N}((\I_k\otimes\X_{ij})\vecop(\mathbf{Y}_{jk}) + \vecop(\mathbf{C}_{ik}), \Sigma_{\mathbf{Z}_k}\otimes\Sigma_{\mathbf{Z}_i})
\end{align}

We recognize the resultant distribution as following into the form $y \sim \mathcal{N}(Mx+b, \Sigma)$. Now, the standard gaussian marginalization identity (e.g. Bishop et al. 2006) can be applied: 

\begin{align}
\vecop(\mathbf{Z}_{ik})\mid\mathbf{X}_{ij} \sim \mathcal{N}((\I_k\otimes\X_{ij})\vecop(\mathbf{B}_{jk}) + \vecop(\mathbf{C}_{ik}), \Sigma_{\mathbf{Z}_k}\otimes\Sigma_{\mathbf{Z}_i} + (\I_k\otimes\X_{ij})(\Sigma_{\mathbf{Y}_k}\otimes\Sigma_{\mathbf{Y}_j})(\I_k\otimes\X_{ij})\trp )
\end{align}

We collect terms using the mixed-product property of kronecker products: 

\begin{align}
\vecop(\mathbf{Z}_{ik})\mid\mathbf{X}_{ij} \sim \mathcal{N}(\vecop(\X_{ij}\mathbf{B}_{jk}) + \vecop(\mathbf{C}_{ik}), \Sigma_{\mathbf{Z}_k}\otimes\Sigma_{\mathbf{Z}_i} + \Sigma_{\mathbf{Y}_k}\otimes \X_{ij}\Sigma_{\mathbf{Y}_j}\X_{ij}\trp)
\end{align}

Now, we can see that the marginal density is a matrix-variate normal only if $\Sigma_{\mathbf{Z}_k}= \Sigma_{\mathbf{Y}_k}$ -- that is, the variable we marginalize over has the same covariance in the dimension we are \emph{not} marginalizing over as the marginal density. Otherwise the density is well-defined but not matrix-normal (see \citet{Stegle2011,Rakitsch2013} for efficient inference in this setting). If we let $\Sigma_k:=\Sigma_{\mathbf{Z}_k}= \Sigma_{\mathbf{Y}_k}$, then we can factor out that term and rewrite the marginal density as a matrix normal: 

\begin{align}
\vecop(\mathbf{Z}_{ik})\mid\mathbf{X}_{ij} \sim \mathcal{N}(\vecop(\X\mathbf{B}_{jk}) + \vecop(\mathbf{C}_{ik}), \Sigma_{k}\otimes\Sigma_{\mathbf{Z}_i} + \Sigma_{_k}\otimes \X\Sigma_{\mathbf{Y}_j}\X\trp)\\
\vecop(\mathbf{Z}_{ik})\mid\mathbf{X}_{ij} \sim \mathcal{N}(\vecop(\X\mathbf{B}_{jk}) + \vecop(\mathbf{C}_{ik}), \Sigma_{k}\otimes(\Sigma_{\mathbf{Z}_i} +\X\Sigma_{\mathbf{Y}_j}\X\trp))\\
\mathbf{Z}_{ik}\mid\mathbf{X}_{ij} \sim \mathcal{MN}(\X\mathbf{B}_{jk} + \mathbf{C}_{ik}, \Sigma_{\mathbf{Z}_i} +\X\Sigma_{\mathbf{Y}_j}\X\trp,\Sigma_{k})
\end{align}

Unlike the multivariate normal case, we can apply the same identity over either $\X$ or $\Y$, since if $\X \sim \mathcal{MN}(M, U, V)$ then $\X\trp \sim \mathcal{MN}(M\trp, V, U)$. We write it directly below: 

\begin{align}
\mathbf{Z\trp}_{ik}\mid\mathbf{X}_{ij},\mathbf{Y}_{jk} \sim \mathcal{MN}(\mathbf{Y}_{jk}\trp\mathbf{X}_{ij}\trp + \mathbf{C}\trp_{ik}, \Sigma_{\mathbf{Z}_k},\Sigma_{\mathbf{Z}_i})\\
\mbox{let } \Sigma_i := \Sigma_{\mathbf{Z}_i}=\Sigma_{\mathbf{X}_i} \\
\cdots\\
\mathbf{Z\trp}_{ik}\mid\mathbf{Y}_{jk} \sim \mathcal{MN}(\mathbf{A}_{jk}\trp\mathbf{X}_{ij}\trp + \mathbf{C}\trp_{ik}, \Sigma_{\mathbf{Z}_k} + \Y\trp\Sigma_{\mathbf{Y}_j}\Y,\Sigma_{\mathbf{Z}_i})\\
\mathbf{Z}_{ik}\mid\mathbf{Y}_{jk} \sim \mathcal{MN}(\mathbf{X}_{ij}\mathbf{A}_{jk}+ \mathbf{C}_{ik},\Sigma_{\mathbf{Z}_i},\Sigma_{\mathbf{Z}_k} + \Y\trp\Sigma_{\mathbf{Y}_j}\Y)
\end{align}

Next, we do the same for the partitioned gaussian identity. First two vectorized matrix-normals that form our partition: 

\begin{align}
\mathbf{X}_{ij} &\sim \mathcal{MN}(\mathbf{A}_{ij}, \Sigma_{i}, \Sigma_{j}) \rightarrow \vecop[\mathbf{X}_{ij}] \sim \mathcal{N}(\vecop[\mathbf{A}_{ij}], \Sigma_{j}\otimes \Sigma_{i})\\
\mathbf{Y}_{ik} &\sim \mathcal{MN}(\mathbf{B}_{ik}, \Sigma_{i}, \Sigma_{k}) \rightarrow \vecop[\mathbf{Y}_{ik}] \sim \mathcal{N}(\vecop[\mathbf{B}_{ik}], \Sigma_{k}\otimes \Sigma_{i})\\
\begin{bmatrix}\vecop[\mathbf{X}_{ij}] \\ \vecop[\mathbf{Y}_{ik}]
\end{bmatrix}
& \sim \mathcal{N}\left(\vecop\begin{bmatrix}\mathbf{A}_{ij} \\ \mathbf{B}_{ik}
\end{bmatrix}
, \begin{bmatrix} \Sigma_{j}\otimes \Sigma_i  & \Sigma_{jk} \otimes \Sigma_i  \\ 
\Sigma_{kj}\otimes \Sigma_i & \Sigma_{k} \otimes \Sigma_i\end{bmatrix}\right)
\end{align}

We apply the standard partitioned Gaussian identity and simplify using the properties of the $\mathrm{vec}$ operator and the mixed product property of kronecker products:

\begin{align}
\vecop[\X_{ij}] \mid \vecop[\Y_{ik}]\sim\mathcal{N}(&\vecop[\A_{ij}] + (\Sigma_{jk}\otimes\Sigma_i)(\Sigma_k\inv\otimes\Sigma_i\inv)(\vecop[\Y_{ik}]-\vecop[\B_{ik}]),\\
 & \Sigma_j\otimes\Sigma_i -  (\Sigma_{jk}\otimes\Sigma_i)(\Sigma_k\inv\otimes\Sigma_i\inv) (\Sigma_{kj}\otimes\Sigma_i))\\
=\mathcal{N}(&\vecop[\A_{ij}] + (\Sigma_{jk}\Sigma_k\inv\otimes\Sigma_i\Sigma_i\inv)(\vecop[\Y_{ik}]-\vecop[\B_{ik}]), \\
 & \Sigma_j\otimes\Sigma_i -  (\Sigma_{jk}\Sigma_k\inv\Sigma_{kj}\otimes\Sigma_i\Sigma_i\inv \Sigma_i))\\
=\mathcal{N}(&\vecop[\A_{ij}] + (\Sigma_{jk}\Sigma_k\inv\otimes\I)(\vecop[\Y_{ik}]-\vecop[\B_{ik}]), \\
 & \Sigma_j\otimes\Sigma_i -  (\Sigma_{jk}\Sigma_k\inv\Sigma_{kj}\otimes\Sigma_i)\\
=\mathcal{N}(&\vecop[\A_{ij}] + \vecop[\Y_{ik}-\B_{ik}\Sigma_k\inv\Sigma_{kj}], (\Sigma_j-\Sigma_{jk}\Sigma_k\inv\Sigma_{kj})\otimes\Sigma_i)
\end{align}

Next, we recognize that this multivariate gaussian is equivalent to the following matrix variate gaussian: 

\begin{align}
\X_{ij} \mid \Y_{ik}\sim \mathcal{MN}(&\A_{ij} +(\Y_{ik}-\B_{ik})\Sigma_k\inv\Sigma_{kj}, \Sigma_i, \Sigma_j-\Sigma_{jk}\Sigma_k\inv\Sigma_{kj})
\end{align}

The conditional in the other direction can be written by working through the same algebra: 

\begin{align}
\Y_{ik} \mid \X_{ij}\sim \mathcal{MN}(&\B_{ik} +(\X_{ij}-\A_{ij})\Sigma_j\inv\Sigma_{jk}, \Sigma_i, \Sigma_k-\Sigma_{kj}\Sigma_j\inv\Sigma_{jk})
\end{align}

Finally, vertical rather than horizontal concatenation (yielding a partitioned row rather than column covariance) can be written by recognizing the behavior of the matrix normal under transposition: 

\begin{align}
\X\trp_{ji} \mid \Y\trp_{ki}\sim \mathcal{MN}(&\A\trp_{ji} +\Sigma_{jk}\Sigma_k\inv(\Y\trp_{ki}-\B\trp_{ki}), \Sigma_j-\Sigma_{jk}\Sigma_k\inv\Sigma_{kj}, \Sigma_i)\\
\Y\trp_{ki} \mid \X\trp_{ji}\sim \mathcal{MN}(&\B\trp_{ki} +\Sigma_{kj}\Sigma_j\inv(\X\trp_{ji}-\A\trp_{ji}), \Sigma_k-\Sigma_{kj}\Sigma_j\inv\Sigma_{jk}, \Sigma_i)
\end{align}

\newpage

\section{ Appendix C : Expectation Conditional Maximization (ECM) derivation for Matrix-Normal Shared Response Model}

The Q function, marginalized $\W$

\begin{align}
&\X \sim \mathcal{MN}(\W\Sb+\bi,\rho\otimes\Sigma_v, \Sigma_t)\\
&\Sb \sim \mathcal{MN}(0,\I,\Sigma_t)\\
&\W \sim \mathcal{MN}(0,\rho\otimes\Sigma_v,\I)
\end{align}

\begin{align}
\mathcal{L} := \expect_{p(\W\mid\X,\theta')}\log p(\X,\W\mid\theta ) =& \frac12 \ew\left[nv\log|\Sigma_t\inv|+ tv\log|\rho\inv| + tn\log|\Sigma_v\inv|\right. \nonumber \\
&- \Tr\left[\Sigma_t\inv(\X-\W\Sb-\bi)\trp(\rho\otimes\Sigma_v)\inv(\X-\W\Sb-\bi)\right]\nonumber\\
&+kv\log|\rho\inv|+kn\log|\Sigma_v\inv| - \Tr\left[\Sigma_{w}\inv\W\trp(\rho\otimes\Sigma_v)\inv\W\right] \nonumber\\
&+\left.k\log|\Sigma_t\inv| -\Tr[\Sigma_t\inv\Sb\trp\Sb]  \right] + \mathrm{const.}_{\theta}\\
=& \frac12 \left[(nv+k)\log|\Sigma_t\inv|+ v(k+t)\log|\rho\inv| + n(k+t)\log|\Sigma_v\inv| \right.\nonumber\\
&- \Tr\left[\Sigma_t\inv(\X-\bi-\W'\Sb)\trp(\rho\otimes\Sigma_v)\inv(\X-\bi-\W'\Sb)\right]\nonumber\\
&- \Tr\left[\Sigma_{w}\inv\W'\trp(\rho\otimes\Sigma_v)\inv\W'\right] -\Tr[\rho\inv\rho']\Tr[\Sigma_v\inv\Sigma_v']\Tr[\Sigma_w'\Sb\trp\Sigma_t\inv\Sb]\nonumber \\
& \left.- \Tr[\rho\inv\rho']\Tr[\Sigma_v\inv\Sigma_v']\Tr[\Sigma_{w}\inv\Sigma_{w}']- \Tr[\Sigma_t\inv\Sb\trp\Sb]  \right] + \mathrm{const.}_{\theta} 
\end{align}

The sufficient statistics are:

\begin{align}
\W\mid\X,\theta \sim & \mathcal{MN}\left(\W', \rho'_w\otimes\Sigma_{vw}',\Sigma_w'\right)\\
{\Sigma}_w'  :=& \I-S(\Sigma_t +\Sb\trp\Sb)\inv\Sb\trp = (\I + \Sb\Sigma_t\inv\Sb\trp)\inv\\
\rho_w' :=& \rho_w\\
\Sigma_{vw}' :=& \Sigma_{vw}\\
\W' =& (\X-\bi)\Sb\trp(\Sigma_t +\Sb\trp\Sb)\inv = (\X-\bi)\Sigma_t\inv\Sb\trp\Sigma_w'
\end{align}

\subsection{Gradients for $\Sb$}

\begin{align}
\d_{\S}\mathcal{L} =& \frac12 \d\left[- \Tr\left[\Sigma_t\inv(\X-\bi-\W'\Sb)\trp(\rho\otimes\Sigma_v)\inv(\X-\bi-\W'\Sb)\right]\right.\\
&\left.-\Tr[\rho\inv\rho_w']\Tr[\Sigma_v\inv\Sigma_{vw}']\Tr[\Sigma_w'\Sb\Sigma_t\inv\S] -\Tr[\Sigma_t\inv\Sb\trp\Sb] \right] \\
=&\frac12 \left[- 2\Tr\left[\Sigma_t\inv(\X-\bi-\W'\Sb)\trp(\rho\otimes\Sigma_v)\inv(\W'\d\Sb)\right]\right.\\
&\left.-2\Tr[\rho\inv\rho_w']\Tr[\Sigma_v\inv\Sigma_{vw}']\Tr[\Sigma_w'\d\Sb\Sigma_t\inv\Sb\trp] \inv\Sigma_{w}'] - 2\Tr[\Sigma_t\inv\Sb\trp\d\Sb] \right] \\
=& -\Tr\left[\Sigma_t\inv(\X-\bi-\W'\Sb)\trp(\rho\otimes\Sigma_v)\inv\W'\d\Sb\right]\\
&-\Tr[\rho\inv\rho_w']\Tr[\Sigma_v\inv\Sigma_{vw}']\Tr[\Sigma_t\inv\Sb\trp\Sigma_w'\d\Sb]- \Tr[\Sigma_t\inv\Sb\trp\d\Sb]\\
\\
\frac{\partial\mathcal{L}}{\partial{\Sb}} =& \W'\trp(\rho\inv\otimes\Sigma_v\inv)(\X-\bi-\W'\Sb)\Sigma_t\inv - \Tr[\rho\inv\rho_w']\Tr[\Sigma_v\inv\Sigma_{vw}']\Sigma_w'\Sb\Sigma_t\inv-\Sb\Sigma_t\inv\\
 =& \W'\trp(\rho\inv\otimes\Sigma_v\inv)(\X-\bi) - (\W'\trp(\rho\inv\otimes\Sigma_v\inv)\W'\Sb - \Tr[\rho\inv\rho_w']\Tr[\Sigma_v\inv\Sigma_{vw}']\Sigma_w'\Sb - \Sb\\
 =& \W'\trp(\rho\inv\otimes\Sigma_v\inv)(\X-\bi) - (\W'\trp(\rho\inv\otimes\Sigma_v\inv)\W' + \Tr[\rho\inv\rho_w']\Tr[\Sigma_v\inv\Sigma_{vw}']\Sigma_w' +1)\Sb\\
\\
\widehat{\Sb} &= (\W'\trp(\rho\inv\otimes\Sigma_v\inv)\W' + \Tr[\rho\inv\rho_w']\Tr[\Sigma_v\inv\Sigma_{vw}']\Sigma_w'+1)\inv\W'\trp(\rho\inv\otimes\Sigma_v\inv)(\X-\bi)
\end{align}

\subsection{Gradients for $\vec{b}$}

\begin{align}
\d_{\vec{b}}\mathcal{L}  =& -\frac12 \Tr\left[\Sigma_t\inv(\X-\bi-\W'\Sb)\trp(\rho\otimes\Sigma_v)\inv(\X-\bi-\W'\Sb)\right]\\
=& - \Tr\left[\vec{1}\trp\Sigma_t\inv(\X-\bi-\W'\Sb)\trp(\rho\otimes\Sigma_v)\inv\d\vec{b}\right]\\
\frac{\partial\mathcal{L}}{\partial\vec{b}} =& (\rho\otimes\Sigma_v)\inv(\X-\bi-\W'\Sb)\Sigma_t\inv\vec{1}\\\\
0=& (\rho\otimes\Sigma_v)\inv(\X-\bi-\W'\Sb)\Sigma_t\inv\vec{1}\\
&\bi\Sigma_t\inv\vec{1}= (\X-\W'\Sb)\Sigma_t\inv\vec{1}\\
&\vec{\hat{b}}= \frac{(\X-\W'\Sb)\Sigma_t\inv\vec{1}}{\sum\Sigma_t\inv}
\end{align}

\subsection{Gradients for $\Sigma_t$}

\begin{align}
\d_{\Sigma_t\inv}\mathcal{L} =& \frac12 \d\left[(nv+k)\log|\Sigma_t\inv|+ v(k+t)\log|\rho\inv| + n(k+t)\log|\Sigma_v\inv| \right.\\
&- \Tr\left[\Sigma_t\inv(\X-\bi-\W'\Sb)\trp(\rho\otimes\Sigma_v)\inv(\X-\bi-\W'\Sb)\right]\\
&- \Tr\left[\Sigma_{w}\inv\W'\trp(\rho\otimes\Sigma_v)\inv\W'\right] -\Tr[\rho\inv\rho']\Tr[\Sigma_v\inv\Sigma_v']\Tr[\Sigma_w'\Sb\Sigma_t\inv\Sb\trp]\\
& \left.- \Tr[\rho\inv\rho']\Tr[\Sigma_v\inv\Sigma_v']\Tr[\Sigma_{w}\inv\Sigma_{w}']- \Tr[\Sigma_t\inv\Sb\trp\Sb]  \right] \\
 =& \frac12 \left[(nv+k)\Tr[\Sigma_t\d\Sigma_t\inv]- \Tr\left[\d\Sigma_t\inv(\X-\bi-\W'\Sb)\trp(\rho\otimes\Sigma_v)\inv(\X-\bi-\W'\Sb)\right]\right.\\
& \left.-\Tr[\rho\inv\rho']\Tr[\Sigma_v\inv\Sigma_v']\Tr[\Sigma_w'\Sb\d\Sigma_t\inv\Sb\trp]- \Tr[\d\Sigma_t\inv\Sb\trp\Sb]  \right] \\
\\
\frac{\partial\mathcal{L}}{\partial\Sigma_t\inv} =&\frac12\left[(nv+k)\Sigma_t - (\X-\bi-\W'\Sb)\trp(\rho\otimes\Sigma_v)\inv(\X-\bi-\W'\Sb)  \right.\\
&\left.-\Tr[\rho\inv\rho']\Tr[\Sigma_v\inv\Sigma_v']\Sb\trp\Sigma_w'\Sb- \Sb\trp\Sb  \right] \\
\\
\widehat{\Sigma_t\inv} =& \left(\frac{1}{nv+k}(\X-\bi-\W'\Sb)\trp(\rho\otimes\Sigma_v)\inv(\X-\bi-\W'\Sb)  \right.\\
&\left.-\Tr[\rho\inv\rho']\Tr[\Sigma_v\inv\Sigma_v']\Sb\trp\Sigma_w'\Sb- \Sb\trp\Sb  \right)\inv
\end{align}

\subsection{Gradients for $\Sigma_v$}

Here again we assume $\rho$ is diagonal, in which case: 

\begin{align}
\d_{\Sigma_v\inv}\mathcal{L}=& \d\frac12 \left[(nv+k)\log|\Sigma_t\inv|+ v(k+t)\log|\rho\inv| + n(k+t)\log|\Sigma_v\inv| \right.\\
&- \Tr\left[\Sigma_t\inv(\X-\bi-\W'\Sb)\trp(\rho\otimes\Sigma_v)\inv(\X-\bi-\W'\Sb)\right]\\
&- \Tr\left[\Sigma_{w}\inv\W'\trp(\rho\otimes\Sigma_v)\inv\W'\right] -\Tr[\rho\inv\rho']\Tr[\Sigma_v\inv\Sigma_v']\Tr[\Sigma_w'\Sb\trp\Sigma_t\inv\Sb]\\
& \left.- \Tr[\rho\inv\rho']\Tr[\Sigma_v\inv\Sigma_v']\Tr[\Sigma_{w}\inv\Sigma_{w}']- \Tr[\Sigma_t\inv\Sb\trp\Sb]  \right] \\
&= \frac12 \left[n(k+t)\Sigma_v\d\Sigma_v\inv \right.\\
&- \sum_j\tau_j\Tr\left[\Sigma_t\inv(\X-\bi-\W'\Sb)\trp\d\Sigma_v\inv(\X-\bi-\W'\Sb)\right]\\
&- \sum_j\tau_j\Tr\left[\Sigma_{w}\inv\W'\trp\d\Sigma_v\inv\W'\right] -\Tr[\rho\inv\rho']\Tr[\Sigma_v'\d\Sigma_v\inv]\Tr[\Sigma_w'(\I+\Sb\trp\Sigma_t\inv\Sb)]\\
\\
\frac{\partial\mathcal{L}}{\partial\Sigma_v\inv} &= \frac12 \left[n(k+t)\Sigma_v \right.\\
&- \sum_j\tau_j(\X-\bi-\W'\Sb)\Sigma_t\inv(\X-\bi-\W'\Sb)\trp\\
&- \left.\sum_j\tau_j\W'\Sigma_{w}\inv\W'\trp -\Tr[\rho\inv\rho']\Tr[\Sigma_w'(\I+\Sb\Sigma_t\inv\Sb\trp)]\Sigma_v'\right]\\
\\
\widehat{\Sigma_v\inv} &= \left(\frac{1}{n(k+t)}\sum_j\tau_j(\X-\bi-\W'\Sb)\Sigma_t\inv(\X-\bi-\W'\Sb)\trp\right.\\
&+\left. \sum_j\tau_j\W'\Sigma_{w}\inv\W'\trp -\Tr[\rho\inv\rho']\Tr[\Sigma_w'(\I+\Sb\Sigma_t\inv\Sb\trp)]\Sigma_v'\right)\inv
\end{align}

\subsection{Constrained covariances}

For template constraints (e.g.\ diagonal, blocked, banded), we can elementwise-multiply the gradient by a template matrix, and construct the constrained update. 

\newpage

\section{ Appendix D : Algorithm for solving kronecker factored matrices}

 In algorithm \ref{alg:kronsolve}, we show how to efficiently solve for a lower triangular matrix that is the kronecker product of smaller lower triangular matrices.
 
 \begin{algorithm}[tb]
    \caption{Solve $x = (L_0 \otimes L_1 \otimes \cdots \otimes L_{n-1}) \backslash y$}
    \label{alg:kronsolve}
 \begin{algorithmic}[1]
    \STATE {\bfseries Input:} vector $y$, matrices $L_0, L_1, \cdots L_{n-1}$
    \STATE {\bfseries Output:} vector $x$ 
    \IF{n == 1}
    \STATE return matrix\_triangular\_solve($L_0, y$)
    \ELSE
    \STATE $x=y$
    \STATE na = dim($L_0$)
    \STATE nb = dim($L_1$)$\times$ dim($L_2$)$\times \cdots$dim($L_{n-1}$)
    \FOR{$i = 0$ {\bfseries to} $na-1$}
    \STATE $t = x[i*nb:(i+1)*nb]/L_0[i,i]$
    \STATE  $x[i*nb:(i+1)*nb]= ( L_1 \otimes \cdots \otimes L_{n-1}) \backslash t$
    \FOR{$j = i+1 $ {\bfseries to} $na-1$}
    \STATE $x[j*nb:(j+1)*nb] -= L_0[j,i] * t$
    \ENDFOR
    \ENDFOR
    \STATE return $x$
    \ENDIF
 \end{algorithmic}
 \end{algorithm}
  Since the cholesky of a kronecker product is the kronecker product of its cholesky factors, we avoid computing the cholesky factorization of a large matrix and instead only cholesky factorize the individual factors. Algorithm \ref{alg:kronsolve} is recursive: line 11 calls the same function but with one less kronecker factor. The masked variant of the algorithm is similar except for lines 4, 11 and 13. Lines 4 and 11 now perform matrix solves with a mask. Line 13 multiplies $L_0[j,i]$ not with $t$ but with $t^\prime = ( L_1 \otimes \cdots \otimes L_{n-1}) \cdot x[i*nb:(i+1)*nb]$. $t$ and $t^\prime$ are identical when no rows and columns are masked, but differ when some of them are masked.
 Solving $\Sigma^{-1}\X$ now involves the following steps - (1) Cholesky factorize the kronecker factor matrices. (2) Use algorithm \ref{alg:kronsolve} to solve $Z = (L_0 \otimes L_1 \otimes \cdots \otimes L_{n-1}) \backslash X$. (3) Apply the corresponding upper triangular variant to solve $(L_0 \otimes L_1 \otimes \cdots \otimes L_{n-1})^T \backslash Z$. 

 We can calculate log-determinant for kronecker products as follows. After cholesky factorization, $\log |\Sigma| = 2\cdot\sum_i((\log|L_i|)(\prod_{j, j\neq i} {\textrm{dim}}(L_j)))$. $\log|L_i|$ is easy to calculate for a triangular matrix $L_i$. For masked kronecker product, the latter product term in the previous expression is replaced by counting the number of valid rows/columns corresponding to that element in the mask.

\newpage

\section{Appendix E : Additional null hypothesis RSA results}

First, we show RSA matrices under the null hypothesis for all subjects and methods: 

\includegraphics{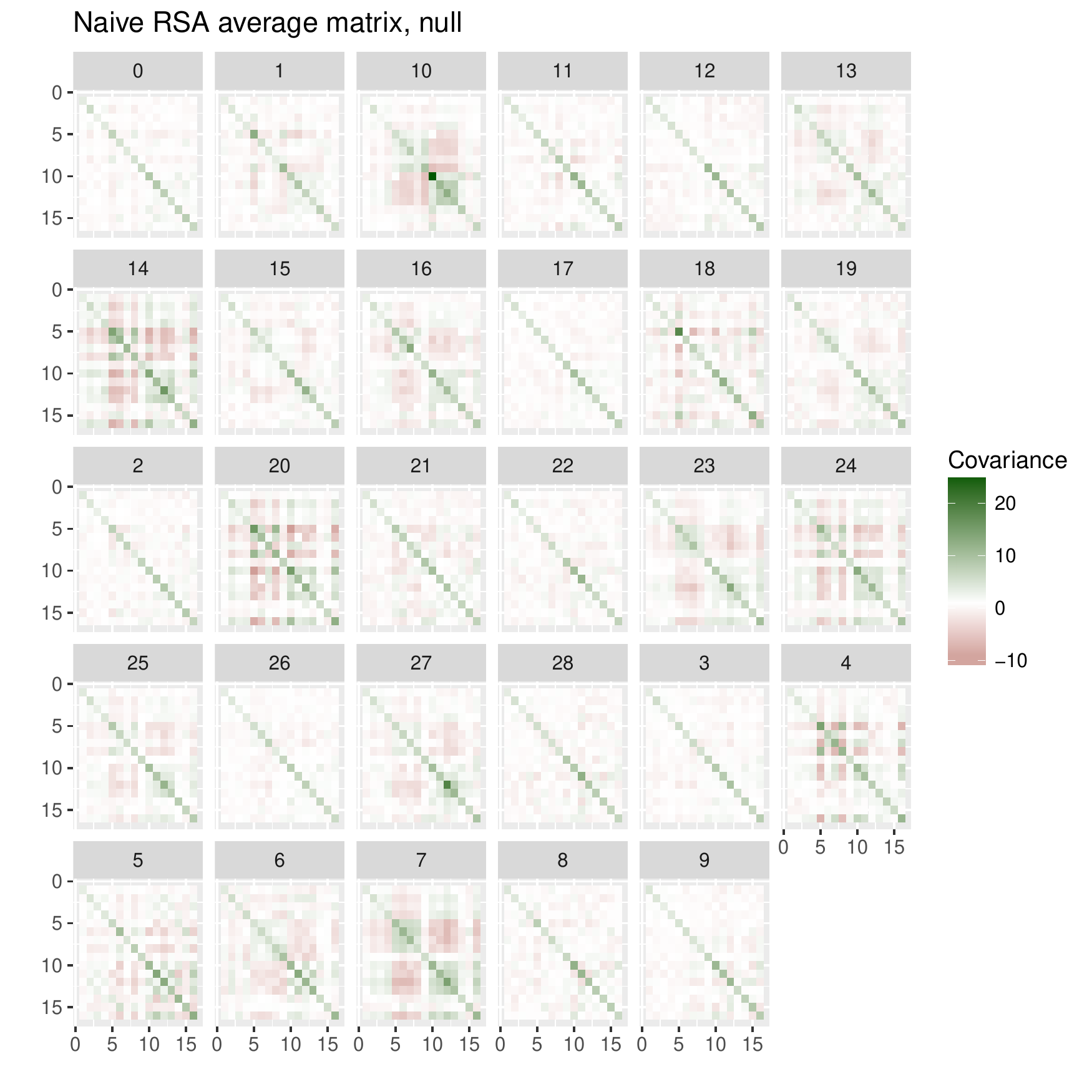}
\newpage

\includegraphics{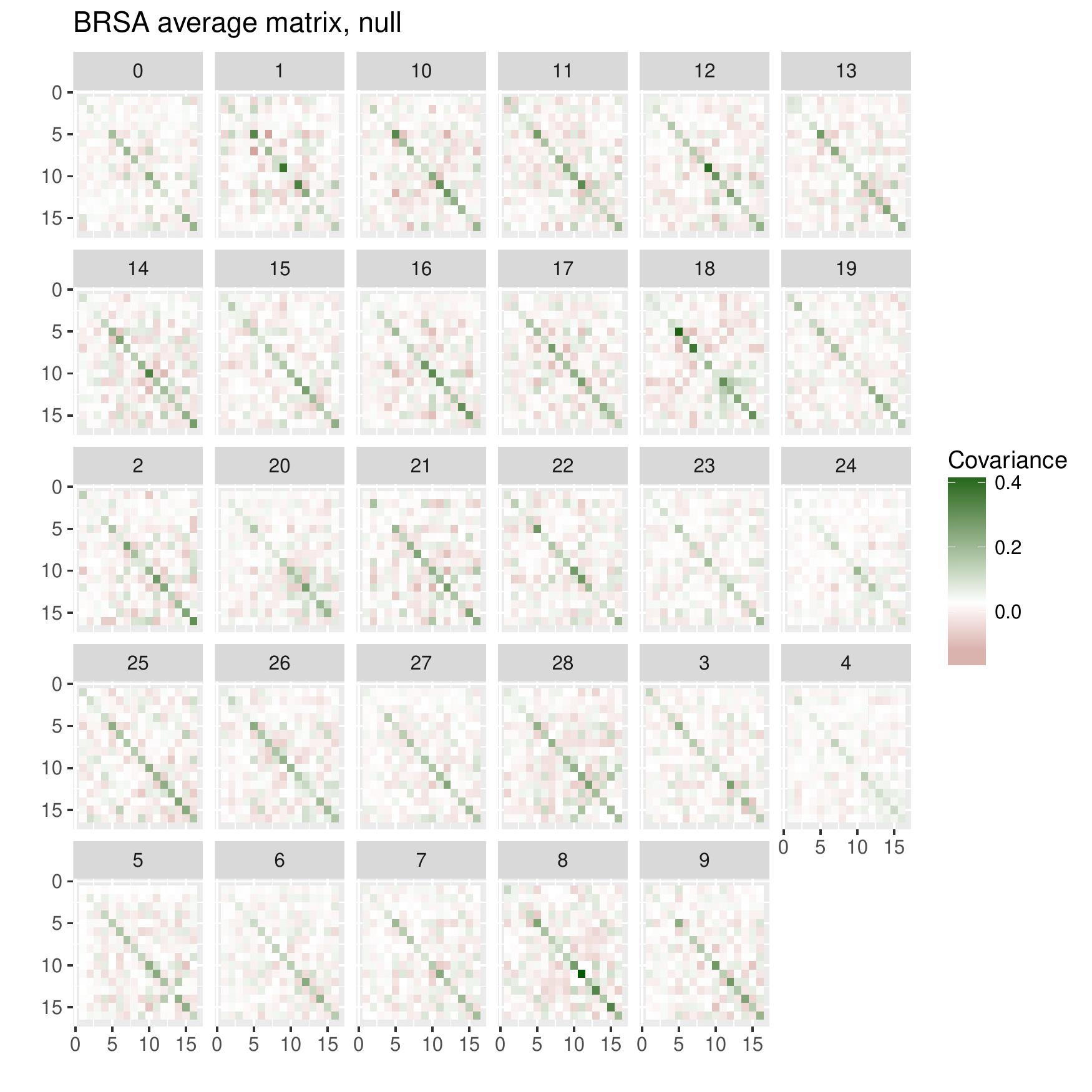}

\newpage
\includegraphics{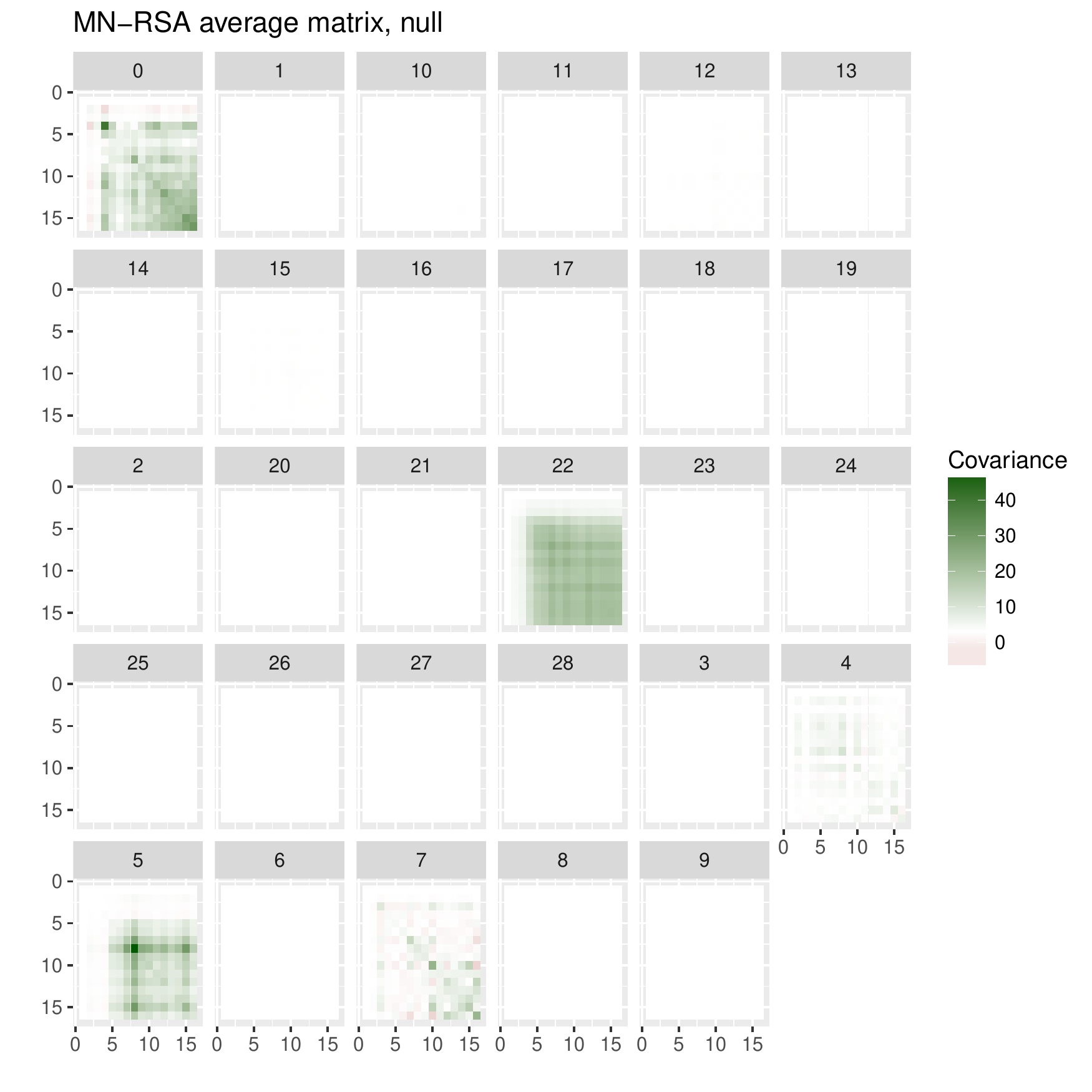}

Notice that only for MN-RSA most of the covariances are noticeably degenerate. This is not a scaling effect on the figure driven by the color bar, but an effect on the underlying data, as we can see in the distribution of values in the covariance matrix for BRSA and MN-RSA: 

\includegraphics[width=0.8\paperwidth]{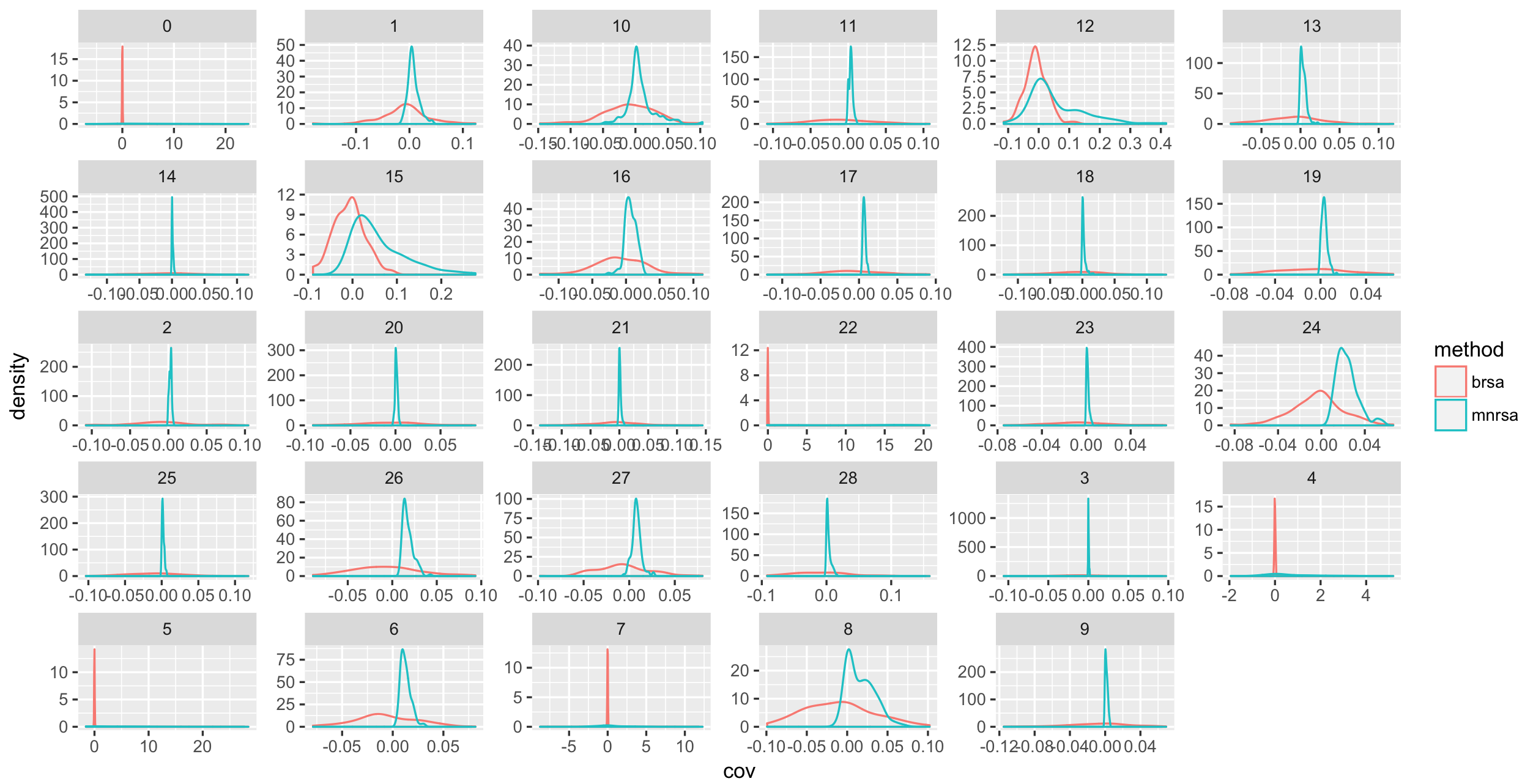}

\section{Appendix F : timing figures for BRSA and MN-RSA}

Experiment details mentioned in main text. Note time on the log scale. 

\includescaled[0.5]{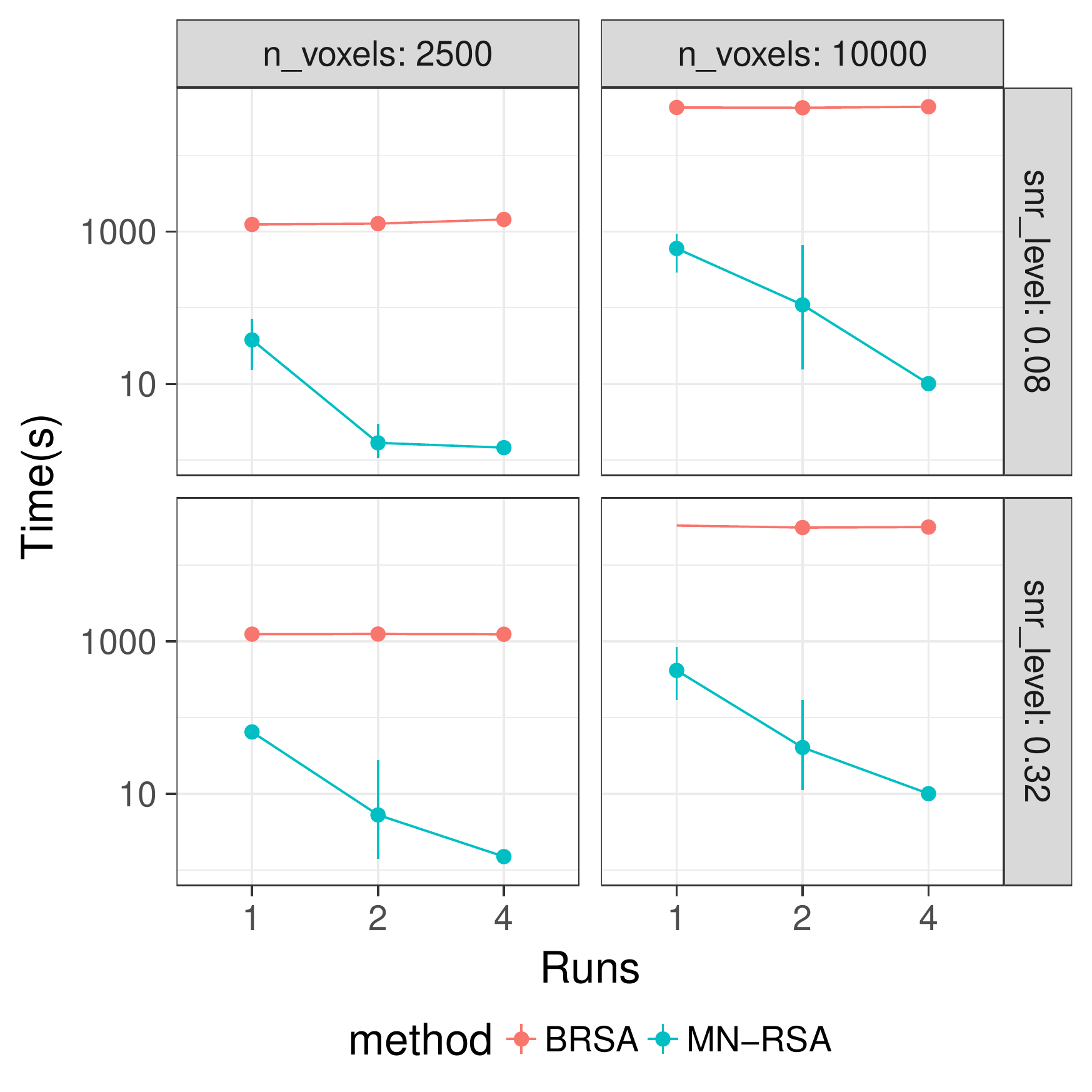}

\end{document}